\documentclass[12pt,twoside]{article}
\usepackage{geometry}
\usepackage[labelfont=bf]{caption}
\usepackage[numbers]{natbib}
\usepackage{amssymb}
\usepackage{graphicx}
\usepackage{float}
\usepackage{booktabs}
\usepackage{tabularx}
\usepackage{titling}
\usepackage{amsmath}
\usepackage{amsthm}
\usepackage{dsfont}
\usepackage{titlesec}
\usepackage{subcaption}
\usepackage{breqn}
\usepackage{thmtools}
\usepackage{url}

\usepackage{algorithm}
\usepackage{algorithmic}

\titlespacing*{\section}
  {0pt}{-.1\baselineskip}{-.1\baselineskip}  
\titlespacing*{\subsection}
   {0pt}{-.1\baselineskip}{-.1\baselineskip}  

\usepackage[T1]{fontenc}

\date{}
\providecommand{\keywords}[1]
{
   \small	
  \textit{\hspace{-1em} Keywords: } #1
}

\title{\textbf{Experimental Comparison of Ensemble Methods and Time-to-Event Analysis Models Through  Integrated Brier Score and Concordance Index}}
\author{\normalsize Camila Fernandez \and \normalsize Chung Shue Chen \and \normalsize Pierre Gaillard \and \normalsize Alonso Silva}

\usepackage{mathtools}

\usepackage{titlesec}
\titleformat{\section}[block]
  {\fontsize{12}{15}\bfseries\sffamily\filcenter}
  {\thesection}
  {1em}
  {\MakeUppercase}
\titleformat{\subsection}[hang]
  {\fontsize{12}{15}\bfseries\sffamily}
  {\thesubsection}
  {1em}
  {}

\begin{document}

\maketitle

\abstract{ \noindent Time-to-event analysis is a branch of statistics that has increased in popularity during the last decades due to its many application fields, such as predictive maintenance, customer churn prediction and population lifetime estimation. In this paper, we review and compare the performance of several prediction models for time-to-event analysis. These consist of semi-parametric and parametric statistical models, in addition to machine learning approaches. Our study is carried out on three datasets and evaluated in two different scores (the integrated Brier score and concordance index). Moreover, we show how ensemble methods, which surprisingly have not yet been much studied in time-to-event analysis, can improve the prediction accuracy and enhance the robustness of the prediction performance. We conclude the analysis with a simulation experiment in which we evaluate the factors influencing the performance ranking of the methods using both scores.\\
}

\section{Introduction}
\label{sec1}

Time-to-event analysis is popular in medical research for predicting the lifetime of populations. It is also widely used in many fields in order    
to predict the time until a certain critical event occurs, which may be the recurrence of a disease, the customer churn in business management and operation research, recidivism in social science and psychology, the failure of machines in industrial engineering, etc. One of the most important characteristics of time-to-event analysis, which makes a significant difference from classical regression problems \cite{zietz2008determinants}, \cite{arik2020tabnet}, is a phenomenon known as censorship, and specifically, in this paper we treat the problem of right censorship. Right censorship arises from the fact that a study may finish before all the samples reach the critical event or because some of the individuals have withdrawn from the study before it ends. As a result, not all the samples may have reached their failure time during the observed period, such that there will be a subset of them whose observed time will represent a lower bound for the critical time.

\medskip
\noindent
Many different models have been proposed in order to predict survival times. One of the most widely used ones was proposed by Cox \cite{cox1972regression} in 1972; this is a semi-parametric model which is composed of an unknown baseline hazard function that depends on the time and the effect of the covariates given by an exponential function. Later, other parametric techniques were proposed by Aalen \cite{aalen1989linear} and Weibull \cite{weibull1939statistical}. Recently, machine learning methods have attracted much attention and many non-parametric model-based machine learning techniques for time-to-event analysis have been proposed, such as random survival forest \cite{ishwaran2008random}, DeepSurv \cite{katzman2018deepsurv}, gradient boosting Cox \cite{ridgeway1999state} and survival support vector machine \cite{polsterl2015fast}. In this paper, we present a comparison of several of these models through two different scores, the concordance index \cite{harrell1996multivariable} and the integrated Brier score  \cite{gerds2006consistent}, and among different types of data sets with the objective to study how the different models behave and compare their effectiveness.

\medskip
\noindent
Ensemble methods are learning algorithms that combine different models by optimizing certain weighting procedures in order to obtain a predictor that will be the combination of multiple learners. One of the main advantages of ensemble methods is the fact that they can inherit the good properties of each of the predictors and use them whenever it is most suitable, for example, if we have a dataset that behaves better for a particular type of models,  then the weighting procedure will privilege this type of models and thus leads to an increment of accuracy that is independent of the chosen dataset. Note that this can be extended to time-varying weighting by which we can also take advantage of the fact that there are some models that vary their performance over time or over the distribution (see \cite{berrisch2021crps}), where we ponder differently the methods that are better for predicting distribution tails and the ones for predicting the center of the distributions. Ensemble methods are well known and used in many applications of data analytics and machine learning, but it is a technique that has not been yet widely explored in time-to-event analysis (see some examples in \cite{zhang2012ensemble}).

\medskip
\noindent
The existing literature lacks clean performance comparison between time-to-event analysis methods and how to calibrate parameters. Van Wieringen et al. \cite{van2009survival} reviewed the performance of different methods applied to the particular case of gene expression data. The methods that are able to handle this type of problem, where the number of features exceeds by far the number of samples, are very specific and do not necessarily represent the general case of survival analysis problems. 

\medskip
\noindent
\textbf{Contributions.} The main contribution of this paper is to give a detailed comparison of different and diverse time-to-event analysis methods using two widely used scores. The above gives us a detailed comparative study of the time-to-event analysis models and their different advantages and disadvantages. To this end, we compare the performance using three datasets and we study the impact of optimizing the hyperparameters through a randomized search. We observe that the method ranking varies across each dataset, making it challenging to select the most appropriate model without prior knowledge. To address this issue, we propose combining these different methods to enhance robustness across datasets. This is carried out by optimizing the parameters of a convex combination of the methods described in Section~\ref{sec3}, such that the integrated Brier score is minimized. Finally, we conduct simulation experiments aimed at deepening insights from the dataset comparison and studying the factors influencing method performance ranking. We generate data using three different techniques under three scenarios: increasing the number of samples, reducing the number of features, and augmenting the percentage of censorship.

\medskip
\noindent
\textbf{Paper outline.} First, we present the preliminaries and definitions for our study, together with the implemented methods: Cox proportional hazard, Gradient boosting Cox, Random survival forest, Weibull accelerated failure time, Aalen's additive and DeepSurv. In Section \ref{sec4}, we exhibit our implementation of ensemble methods. In Section~\ref{sec5}, we present the three datasets (Primary biliary cirrhosis, German breast cancer and Telecom churn) used for our study. Section~\ref{sec6} shows the comparison of the various techniques and their numerical results. Besides, we show the performance of the ensemble method. In Section \ref{sec:simexp}, we present the simulation experiment and finally, Section \ref{sec7} contains some concluding remarks.\\

\section{Preliminaries}
\label{sec2}

The main objective of time-to-event analysis is to estimate the distribution of survival times. Given a set of $N$ subjects with its respective vector of covariates of dimension $d$, $x_i = \{x_i^1, \ldots, x_i^d  \} \in \mathcal{X}, ~ i \in \{1,\ldots,N\} $, we assume that $x_i$ follows the distribution of a random variable $X_i$. Let $T_i$ and $C_i$ be a non-negative random variable denoting the survival and censored time, respectively. Then, we define the observed time as $Y_i = \min \{T_i, C_i\}$ and we will write $\Delta_i = \mathds{1}\{ T_i \leq C_i\}$ for the survival indicator. Under these conditions, a subject of the dataset will be described by $(x_i,y_i, \delta_i) \in \mathcal{X} \times \mathbb{R} \times \{0,1\}$ assumed to be a realization of the random variable $(X_i, Y_i, \Delta_i)$. In addition, we define the set of individuals at risk as $\mathcal{R}(t) = \{ i \in \{1,\ldots, N\}: y_i > t\}$. Let us remark that we consider that all the individuals are present at time $t=0$. Then, the probability to survive at time $t$ for subject $i$ of the dataset is given by:
\begin{equation} S(t\vert x_i) =  \mathbb{P}(T_i > t \vert X_i = x_i). \nonumber 
\end{equation}

\noindent
In order to estimate the survival probability, many parametric and semi-parametric models assume a particular shape of the hazard function, which is defined for all $t>0$ as:
\begin{align}
h(t\vert x_i) = & - \frac{\partial}{\partial t} \log( S(t\vert x_i))\nonumber  \\
=  & \lim_{dt \rightarrow 0} \frac{ \mathbb{P} (t \leq T_i < t + dt\vert T_i \geq t, X_i = x_i)}{dt}. \nonumber
\end{align}
We can retrieve the survival probability function by integrating the exponential of the hazard function 
\begin{equation}
S(t\vert x_i) = \exp \left( - \int\limits_{0}^{t} h(u\vert x_i)du \right). \nonumber
\end{equation}

\noindent
Each model will give us an estimator $\hat{S}$ of the survival probability $S$. In addition, we define the mortality risk of an individual by a function $R:\mathcal{X} \to \mathbb{R}_+$, which will be used later to compute the concordance index. The mortality risk must satisfy $R(x_i) > R(x_j)$ if $\mathbb{P}(T_i < T_j) > 1/2$, i.e. if individual $i$ has a higher mortality risk than individual $j$. Note that $R$ is not uniquely defined and only the ranking matters. Each model will define and estimate (by providing a function $\hat R$) the mortality risk differently and we give the details separately in Section \ref{App2}. In addition, to measure the goodness of fit of each model, we consider two scores. Concordance Index~\cite{harrell1996multivariable} is a rank score that measures the ability of the model to correctly provide a reliable ranking of the survival times. And secondly, the integrated Brier score~\cite{gerds2006consistent}, which measures the calibration of the models by averaging the square distances between the observed survival status and the predicted survival probability. We give more details about both scores in Section~\ref{App1}.

\medskip
\subsection{Methods and their implementation}
\label{sec3}

We consider six methods in our study. These are Cox proportional hazard \cite{cox1972regression}, gradient boosting Cox \cite{ridgeway1999state}, random survival forest \cite{ishwaran2008random}, Weibull AFT \cite{weibull1939statistical}, Aalen additive \cite{aalen1989linear} and DeepSurv \cite{katzman2018deepsurv}. There exist many other methods for survival analysis, such as life tables \cite{cutler1958maximum}, different versions of cox regressions \cite{binder2008allowing}, \cite{hastie2009elements}, linear regressions \cite{tobin1958estimation}, Bayesian network classifier based methods \cite{friedman1997bayesian} and support vector machine \cite{khan2008support}, see \cite{wang2019machine} for more details. Nevertheless, we choose the six methods mentioned above because they are the most popular and widely used techniques, they include parametric, semi-parametric and machine learning approaches, and on the other hand, the diversity of their structure is very relevant and has a key role in ensemble methods. Note that in our implementation and the comparative study, we adopted the methods from the standard libraries: Scikit-survival \cite{polsterl2020scikit}, Lifelines \cite{davidson2019lifelines} and PySurvival \cite{pysurvival_cite}. More details about the methods can be found in Section~\ref{App2}.\\

\section{Ensemble Methods}
\label{sec4}

The main objective of ensemble methods is to combine the predictions of multiple estimators in order to improve generalizability and robustness and to obtain more reliable and accurate predictions. One has to derive effective combination rules or design powerful algorithms to boost performance. Ensemble methods consist of both empirical \cite{hansen1990neural} and theoretical~\cite{schapire1990strength} approaches. It can be proved that weak learners can be boosted into strong learners through ensemble methods by combining multiple estimators. Applications of ensemble methods \cite{zhou2019ensemble} can be found in many fields, such as computer vision, computer security, aided medical diagnosis, credit card fraud detection, weather forecasting, predictive maintenance, etc.

\medskip
\noindent
There are three reasons why it is possible to construct very good ensemble methods~\cite{dietterich2000ensemble}. First, from a statistical point of view, a learner algorithm can be seen as a procedure to identify the best hypothesis space $\mathcal{H}$. When there is a small amount of data available, the algorithm may find many spaces that fit with the same accuracy. By aggregation, ensemble methods, however, can reduce the risk of choosing the wrong learner. Secondly, ensemble methods have computational advantages because learning algorithms can get stuck in local optimum solutions, and even when there is enough training data, it can still be challenging to find the best hypothesis. This issue can be addressed by running multiple learners from different starting points. Thirdly, in most applications of machine learning, the truth cannot be represented by any of the hypotheses in the $\mathcal{H}$ space. However, by forming a weighted version of the elements of $\mathcal{H}$, it is possible to expand the space of representable functions.               

\medskip
\noindent
In this paper, we use a gradient descent optimization algorithm to set the parameters of the convex combination of the six methods described in Section~\ref{sec3}. Assuming that we have $K$ procedures to estimate the survival probability function, let us set $\hat{S_k}$ as the estimator proposed by the $k$-th method. We want to find the parameters $\lambda_k \geq 0$ such that
\begin{equation}
 \hat{S}(t\vert X) = \sum\limits_{k=1}^K \lambda_k \hat{S_k}(t\vert X), \nonumber
\end{equation}
minimizes the integrated Brier score provided that $\sum \lambda_k = 1$. In order to do this, we optimize the weights $\lambda_k$ in a subset of the data $\mathcal{D}$ of size $n$. We consider the gradient vector as the descent direction, which follows the definition of integrated Brier score 
(see Section \ref{subsec22}), the $j$-partial derivative is given by:
\begin{align}
&\frac{\partial IBS(\hat{S},\mathcal{D})}{\partial \lambda_j}  = \frac{1}{\tau n}  \sum\limits_{i=1}^n \int\limits_0^{\tau} W_i(t) \cdot 2 \big( \mathds{1}\{y_i > t\} \nonumber \\
&- \sum\limits_{k=1}^K \lambda_k \hat{S_k}(t\vert x_i) \big)\cdot\left( - S_j(t\vert x_i)\right)~dt. \label{eq:gradient_def}
\end{align}
The gradient descent algorithm is presented in  Algorithm~\ref{alg:EG}.

\begin{algorithm}
\caption{Exponential Gradient Descent}\label{alg:EG}
\begin{algorithmic}[1]
\REQUIRE  $T$ number of iteration, $\eta >0$ learning rate
\STATE {\bfseries Initialization:} $\lambda(0) = (1/K,\dots,1/K)$
\FOR{each iteration $t=1, \dots, T$}
        \STATE Define $Z_t = \sum_{k=1}^K \lambda_k(t)\exp(-\eta Df_k)$, 
         \STATE where $Df_k = \frac{\partial IBS(\hat{S},\mathcal{D})}{\partial \lambda_k}$ defined in~\eqref{eq:gradient_def}. 
        \STATE Update $\lambda_{k}(t+1) = \frac{\lambda_k(t) \exp(-\eta Df_k)}{Z_t}$ for all $k=1,\dots,K$.
\ENDFOR
\end{algorithmic}
\end{algorithm}

\noindent
Here, we consider $\eta$ a constant learning rate with initial $\lambda$ equitably distributed. The iteration process is repeated until it reaches a maximum number that is set as $10000$. We estimate the optimal aggregation weights each time when we fit the methods in a cross-validation process of five folds. It is important to mention that using a gradient descent algorithm for optimizing the parameters is possible thanks to the fact that the integrated Brier score function is convex, which is not the case for the concordance index.\\ 

\section{Datasets}
\label{sec5}

We study three different datasets, whose general properties are summarized in Table~\ref{tab1}.

\begin{table*}[ht]
\centering
\caption{Dataset dimensions}\label{tab1}
\begin{tabular}{@{}lllll@{}}
\toprule
& Samples  & Features & Censored & Percentage\\
\midrule
PBC \cite{pbcDataset}  & 276   & 17  & 165  & 59.8 \% \\
GBCSG2 \cite{schumacher1994randomized}   & 686   & 8 & 387  & 56.4 \% \\
TLCM \cite{tlcmDataset}  & 7043  & 19  & 5174 & 73 \% \\
\end{tabular}
\end{table*}

\subsection{Primary Biliary Cirrhosis (PBC)}
\label{subsec51}
Mayo Clinic Primary Biliary Cirrhosis dataset was made available by Therneau and Grambsch \cite{pbcDataset} and it is for studying the effects of the drug D-penicillamine on the lifetime of patients with PBC. This dataset has 276 samples and 17 covariates such as age, presence of ascites, cholesterol, etc. There are 165 patients who did not die at the end of the study (59.8\%) and that corresponds to censored data.

\subsection{German Breast Cancer Study Group 2 (GBCSG2)}
\label{subsec52}
German Breast Cancer Study Group was made available by Schumacher et al. \cite{schumacher1994randomized} and it is used for studying the effects of hormone treatment on breast cancer recurrence. The dataset has 686 samples and 8 covariates, such as age, hormonal therapy, menopausal status, etc. There are 387 patients who did not get cancer again (56.4\%), corresponding to censored data.

\subsection{Kaggle Telco Churn (TLCM)}
\label{subsec53}
Kaggle Telco Churn dataset was made available in 2008 by Kaggle and it is a sample dataset from IBM  \cite{tlcmDataset}. It is used for studying the different causes of customer churn in a fictional telecommunication enterprise. The dataset has 7043 samples and 19 features such as gender, partner, dependents, phone service, etc. This dataset has 5174 clients who have not churned at the end of the study (73\%) and that corresponds to censored data.\\

\section{Comparison Results}
\label{sec6}

In the following section, we compare the six methods described in Section \ref{sec3} through concordance index and integrated Brier score, respectively. Besides, we compare their results with that of the deployed ensemble method. For each dataset, the scores were computed 25 different times corresponding to 25 partitions (training/validation) of the dataset. This number was chosen arbitrarily in order to maintain a reasonable number of iterations without making the process too computationally expensive. Results are shown by the box plots below. Note that among Figure \ref{fig:test1} to \ref{fig:test6}, there are some methods with their names marked with an asterisk and their boxes colored by red, which is to indicate the implementation of a randomized search of the parameters conducted by a cross-validation process, whereas the unmarked (and blue) corresponds to adjust the method with the default parameters described in Section \ref{sec3}. In addition, the machine learning techniques were bolded to differentiate them from the semi-parametric and parametric methods. 

\subsection{Concordance index comparison}
\label{subsec61}

\begin{figure}[H]
\centering
\begin{subfigure}[t]{0.49\textwidth}
\includegraphics[width=\linewidth]{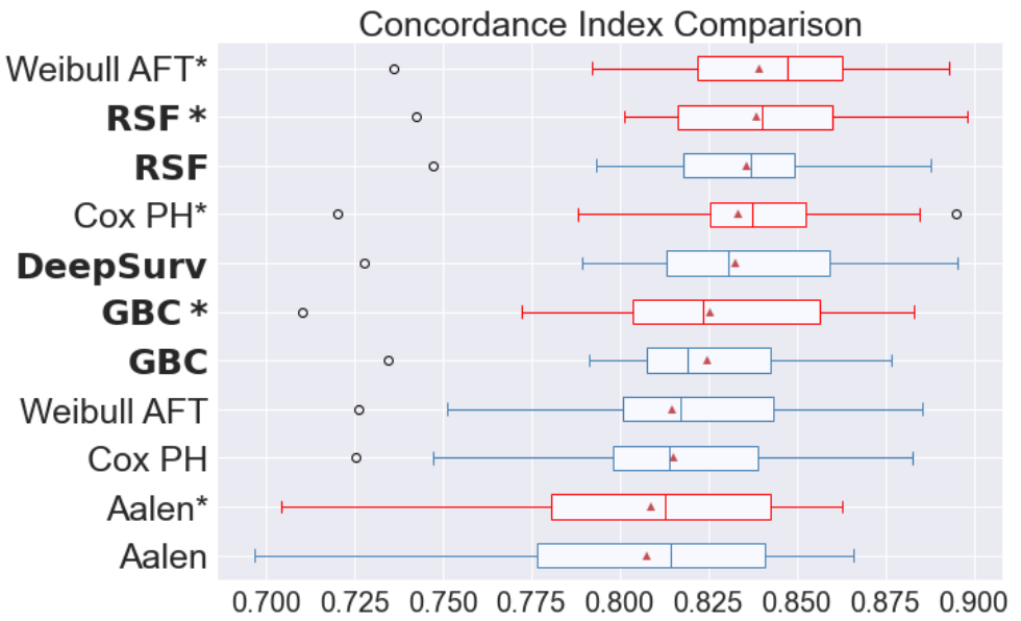}
  \caption{Concordance index comparison of primary biliary chirrosis dataset}
  \label{fig:test1}
\end{subfigure}
\hfill
\begin{subfigure}[t]{0.49\textwidth}
\includegraphics[width=\linewidth]{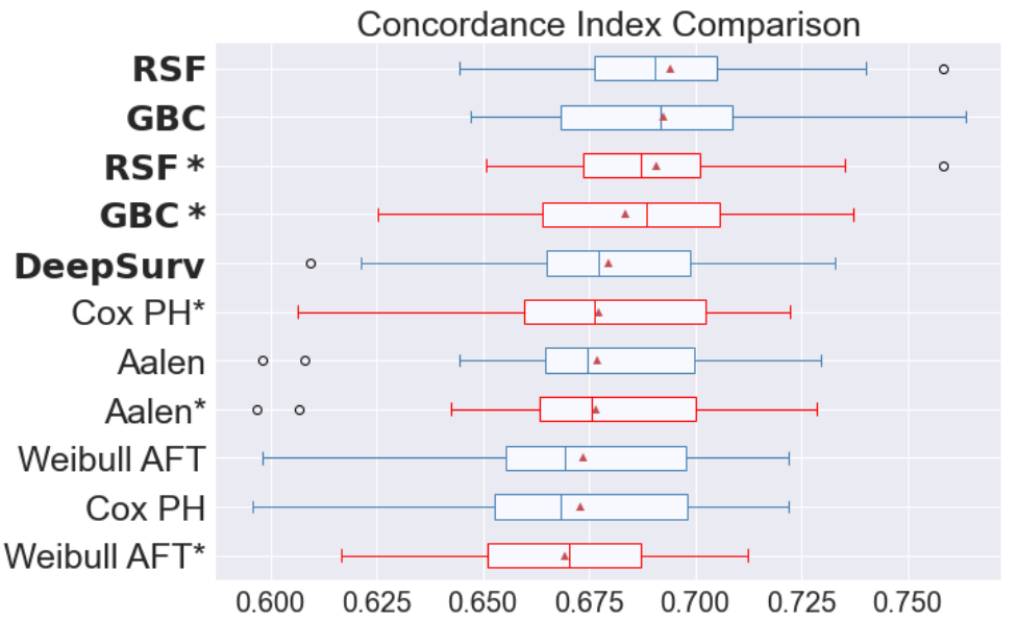}
 \caption{Concordance index comparison of German breast cancer dataset}
 \label{fig:test2}
\end{subfigure}
\caption{}
\end{figure}

\begin{figure}[H]
\centering
 \includegraphics[width = 0.49\linewidth]{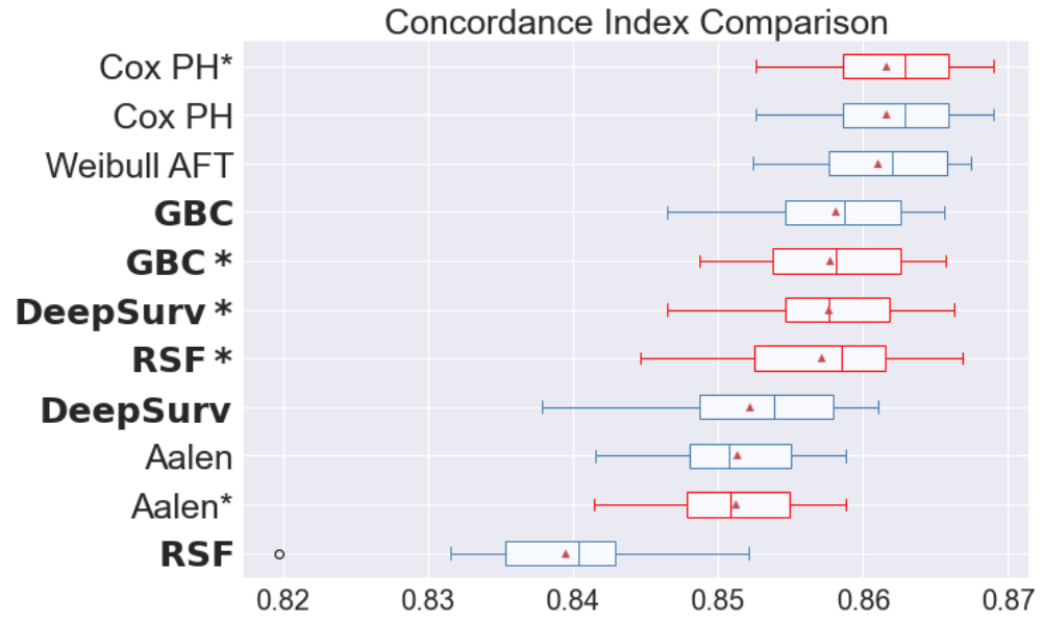}
  \caption{Concordance index comparison of telecom churn dataset}
  \label{fig:test3}
\end{figure}

Figure \ref{fig:test1} shows the concordance index comparison under the PBC dataset. The methods are shown in decreasing order of their obtained mean score. Note that the mean score value is marked by the red triangle in each box plot. We can observe that Weibull AFT with the randomized search of the parameters (denoted by Weibull AFT$^*$) is the method that outperforms the others, followed by random survival forest with the randomized search of the parameters (RSF$^*$), random survival forest (RSF) and Cox proportional hazard with the randomized search of the parameters (Cox PH$^*$). We can also see that the randomized search of the parameters works well for all the methods (see Weibull AFT$^*$ vs. Weibull AFT, RSF$^*$ vs. RSF, Cox PH$^*$ vs. Cox PH, GBC$^*$ vs. GBC, and Aalen$^*$ vs. Aalen, respectively). In particular, Weibull AFT$^*$ and Cox PH$^*$ obtain an increment of $2.9\%$ and $2.5\%$ against Weibull AFT and Cox PH, respectively.

\medskip
\noindent
Figure \ref{fig:test2} shows the concordance index comparison result under the GBCSG2 dataset. Here, the method with the best performance is the random survival forest (RSF), followed by gradient boosting Cox (GBC). Unlike the result under the PBC dataset, we cannot observe an increment in the performance when implementing the randomized search of the parameters on RSF, GBC and the other, except for Cox proportional hazard, implementing the randomized search of the parameters (i.e., Cox PH$^*$) has a slight increment of $0.7\%$. Figure \ref{fig:test3} shows the concordance index comparison result under the TLCM dataset. We see that the Cox proportional hazard method (both Cox PH$^*$ and Cox PH) outperforms the others, followed by Weibull AFT, whose performance is close to Cox's. In this dataset, we observe that the randomized search does not contribute significantly to improving the performance of the methods, except for the case of random survival forest (RSF) where there is a $2 \%$ increment by RSF$^*$ when compared with RSF. Weibull AFT$^*$ and DeepSurv* were not considered in the graph because they underperformed compared to the other models, and in addition, their performance value was out of the bounds of the figure.

\medskip
\noindent
In general, we can observe that for the first two datasets (PBC and GBCSG2), machine learning methods (RSF, RSF*, GBC, GBC* and DeepSurv) perform very well, while parametric methods (Cox PH, Weibull AFT and Aalen additive) are left behind.

\medskip
\noindent
This is not the case for the TLCM dataset where Cox HP and Weibull AFT are leading. In addition, we would like to remark the fact that the performance of each method, and its ranking, depends on the dataset. Some methods will perform better for a certain type of dataset than others, this might be due to the assumptions about the hazard function structure and how these assumptions fit the real distribution of each dataset.   

\subsection{Integrated Brier score comparison}
\label{subsec62}

\begin{figure}[H]
\centering
\begin{subfigure}[t]{0.49\textwidth}
\includegraphics[width=\linewidth]{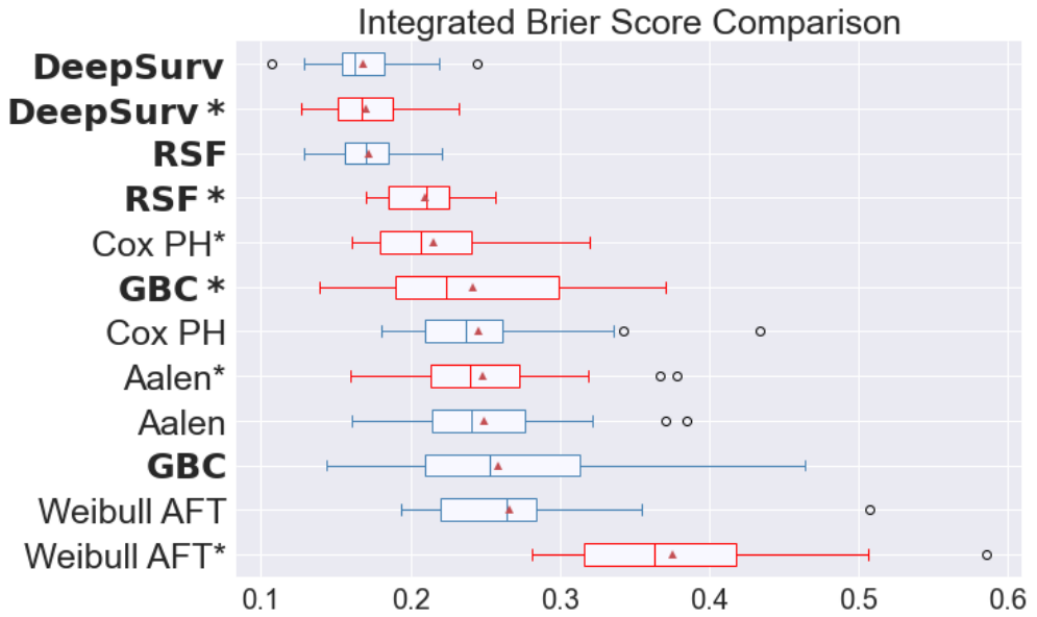}
  \caption{Integrated Brier score comparison of primary biliary chirrosis dataset}
  \label{fig:test4}
\end{subfigure}
\hfill
\begin{subfigure}[t]{0.49\textwidth}
\includegraphics[width=\linewidth]{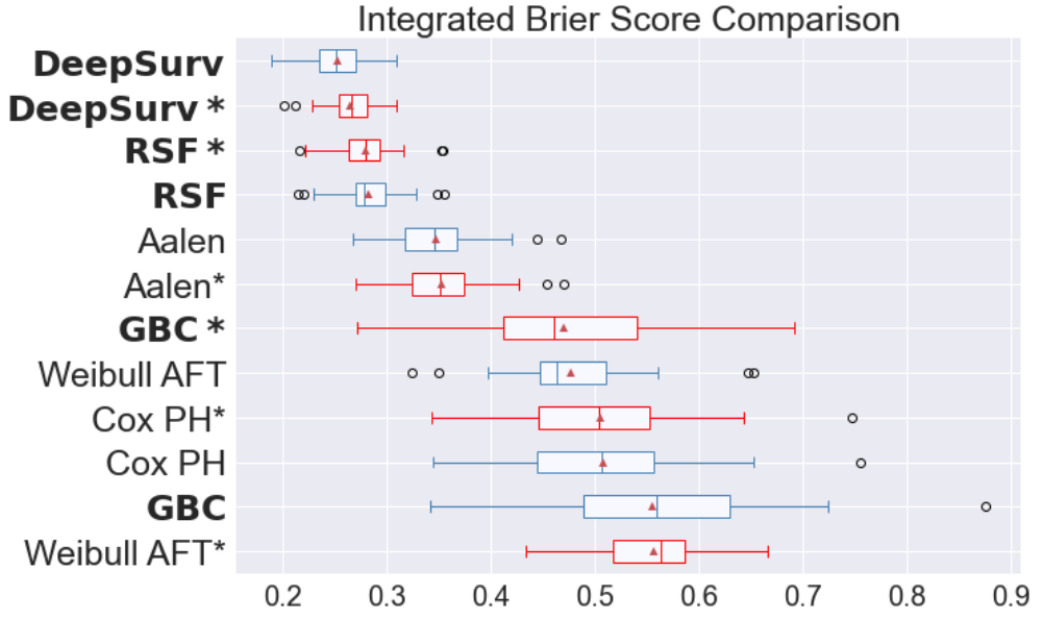}
 \caption{Integrated Brier score comparison of German breast cancer dataset}
 \label{fig:test5}
\end{subfigure}
\caption{}
\end{figure}

\begin{figure}[H]
\centering
 \includegraphics[width = 0.49\linewidth]{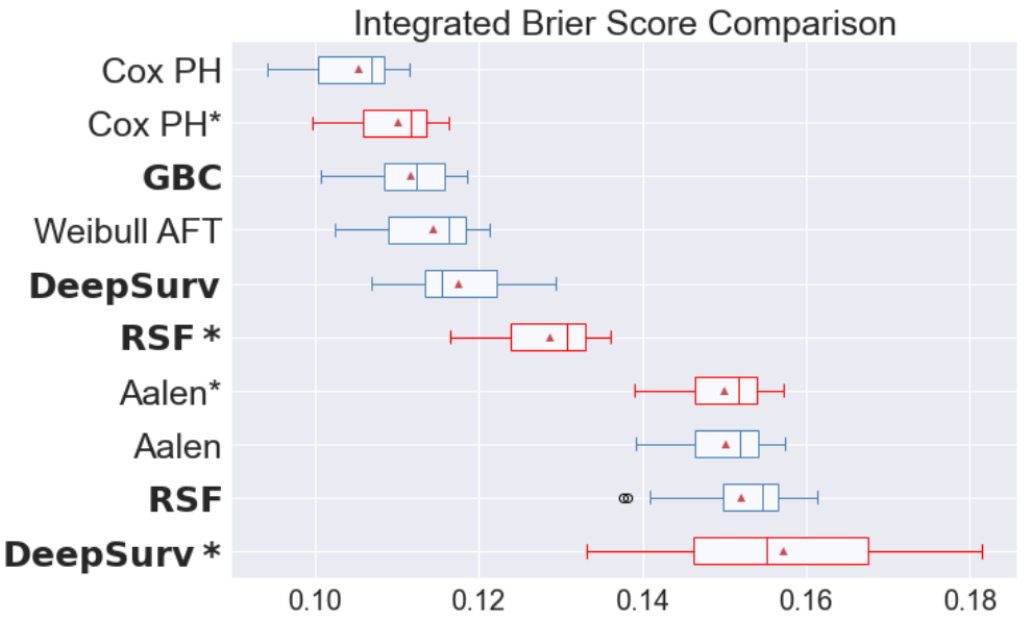}
  \caption{Integrated Brier score comparison of Telecom churn dataset}
  \label{fig:test6}
\end{figure}

Figure \ref{fig:test4} shows the integrated Brier score comparison under the PBC dataset. In this figure, as in the case of the concordance index, the methods are displayed in the increasing order of performance, which in this case corresponds to decreasing integrated Brier score.  Here, we observe that DeepSurv outperforms the other methods (for the IBS score, the lower the better), followed by DeepSurv$^*$ and RSF. We see that there is a clear predominance of machine learning techniques (DeepSurv, DeepSurv*, RSF and RSF*). Similarly, for the GBCSG2 dataset, in Figure \ref{fig:test5}, DeepSurv outperforms the other methods, followed by DeepSurv$^*$, RSF$^*$, and RSF. Note that Aalen additive has a performance of $23\%$ worse than that of RSF. In this case, we can also say that machine learning techniques (DeepSurv, DeepSurv*, RSF and RSF*) have better results than the other methods.

\medskip
\noindent
Figure \ref{fig:test6} shows the integrated Brier score comparison under the TLCM dataset. Contrary to the previous cases, Cox PH method is the lead. In Figure~\ref{fig:test6}, we can appreciate a slight predominance of parametric approaches (Cox PH, Cox PH* and Weibull AFT). We can see that when the amount of censored data is larger, machine learning techniques (DeepSurv and RSF) do not outperform the classical parametric methods.

\medskip
\noindent
Finally, we would like to remark that for a given dataset the results for the concordance index and integrated Brier score differ. This is not surprising in this case due to the nature of the two scores, that is very different in between them. Some models can give good ranked results while calibrating very poorly and vice-versa. Discussions about how to choose an appropriate score have taken place in the past and there is no consensus in the community \cite{van2009survival}.

\subsection{Ensemble methods comparison}
\label{subsec63}

In the following, we show the result of our deployed ensemble method. Each aggregation is set according to Section \ref{sec4} for optimizing the parameters of a convex combination of the six methods (described in Section \ref{sec3}).

\begin{figure}[H]
\centering
\begin{subfigure}[t]{0.49\textwidth}
\includegraphics[width=\linewidth]{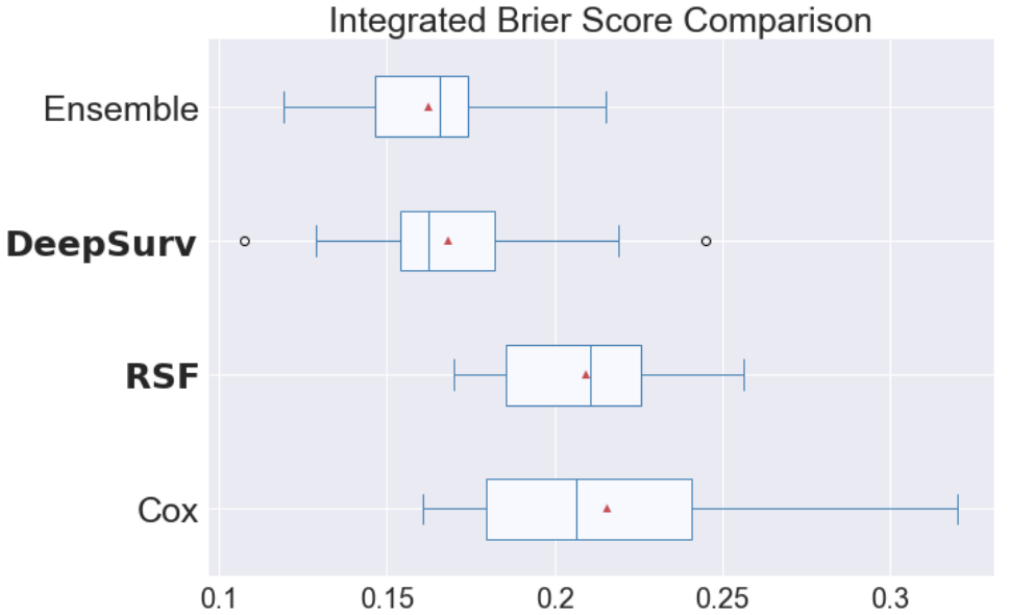}
  \caption{Ensemble method comparison using integrated Brier score on the primary biliary cirrhosis dataset}
  \label{fig:test7}
\end{subfigure}
\hfill
\begin{subfigure}[t]{0.49\textwidth}
\includegraphics[width=\linewidth]{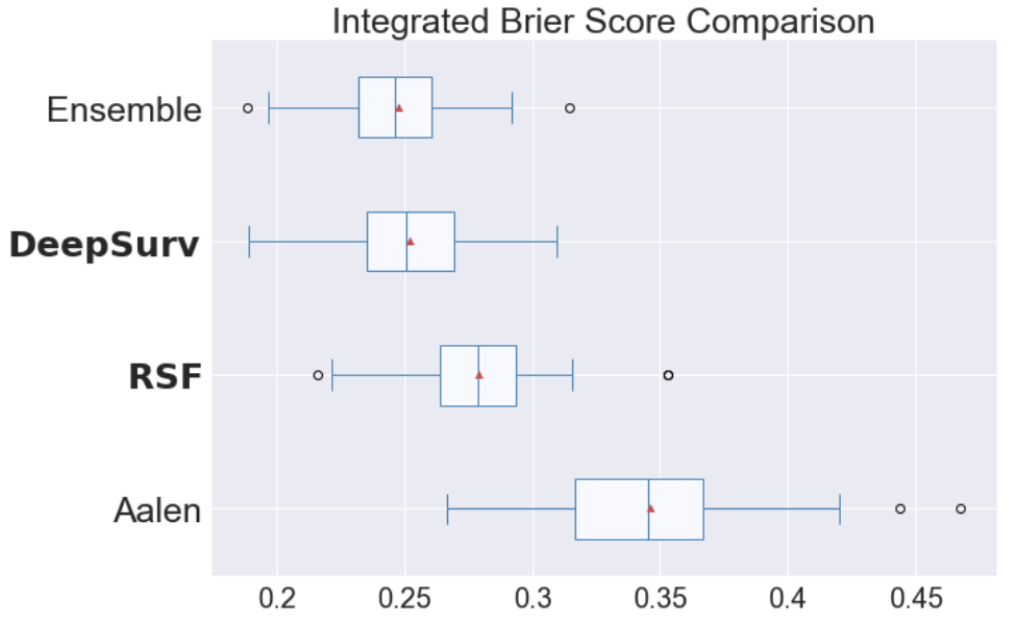}
 \caption{Ensemble method comparison using integrated Brier score on the German breast cancer dataset}
 \label{fig:test8}
\end{subfigure}
\caption{}
\end{figure}

\begin{figure}[H]
\centering
 \includegraphics[width = 0.49\linewidth]{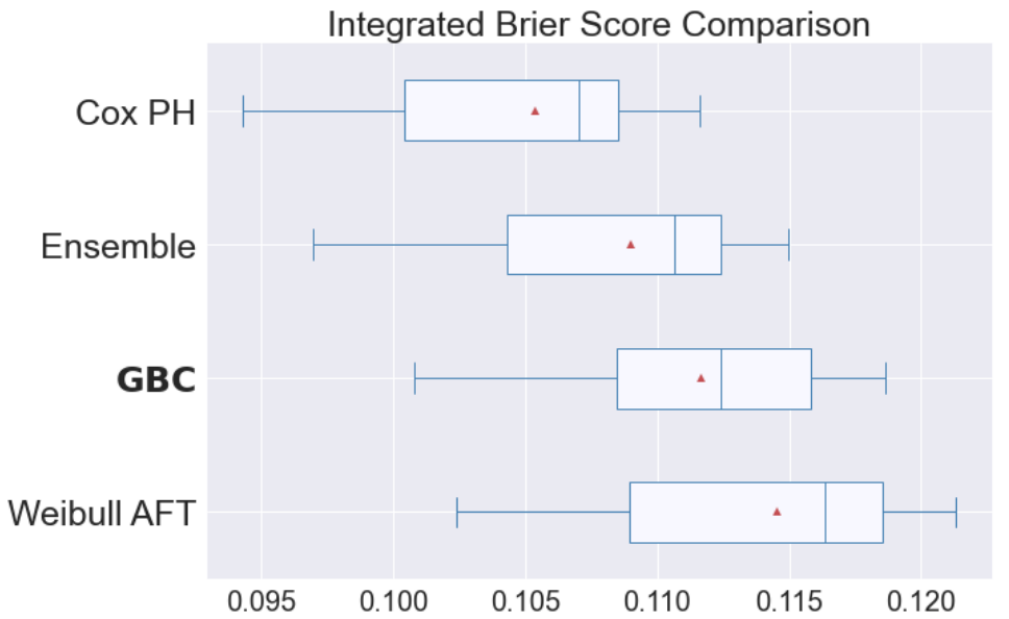}
  \caption{Ensemble method comparison using integrated Brier score on the Telecom churn dataset}
  \label{fig:test9}
\end{figure}

\noindent
Figure \ref{fig:test7} shows the integrated Brier score comparison result under the PBC dataset. We observe that the ensemble method through gradient descent outperforms DeepSurv by $3\%$. Similarly, in Figure \ref{fig:test8} for the GBCSG2 dataset, we find that the ensemble method outperforms the best predictor among the six and obtains a performance improvement of $1.6\%$. Finally, Figure \ref{fig:test9} shows the integrated Brier score comparison result under the TLCM dataset. The ensemble method does not improve the performance, whereas the best estimator is the Cox PH which has a performance of $3.8\%$ better than that of the ensemble method.

\medskip
\noindent
In addition, we show the overall performance of each method by averaging the scores obtained by each under the three datasets so as to estimate their overall performance. 

\begin{figure}[H]
\centering
\begin{subfigure}[t]{0.49\textwidth}
\includegraphics[width=\linewidth]{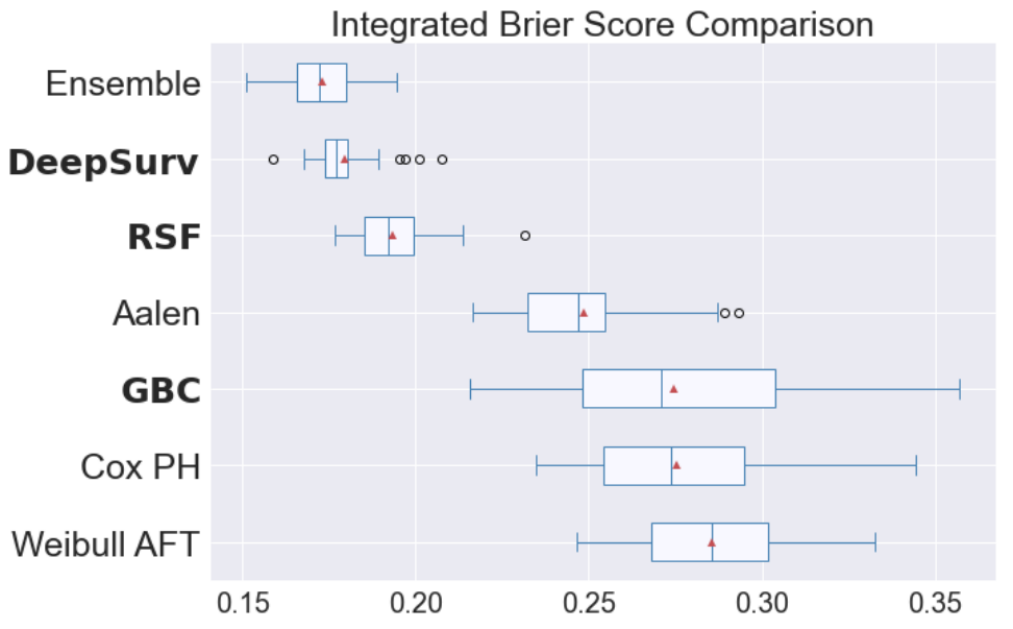}
  \caption{Integrated Brier score comparison among the three datasets} 
  \label{fig:test10}
\end{subfigure}
\hfill
\begin{subfigure}[t]{0.49\textwidth}
\includegraphics[width=\linewidth]{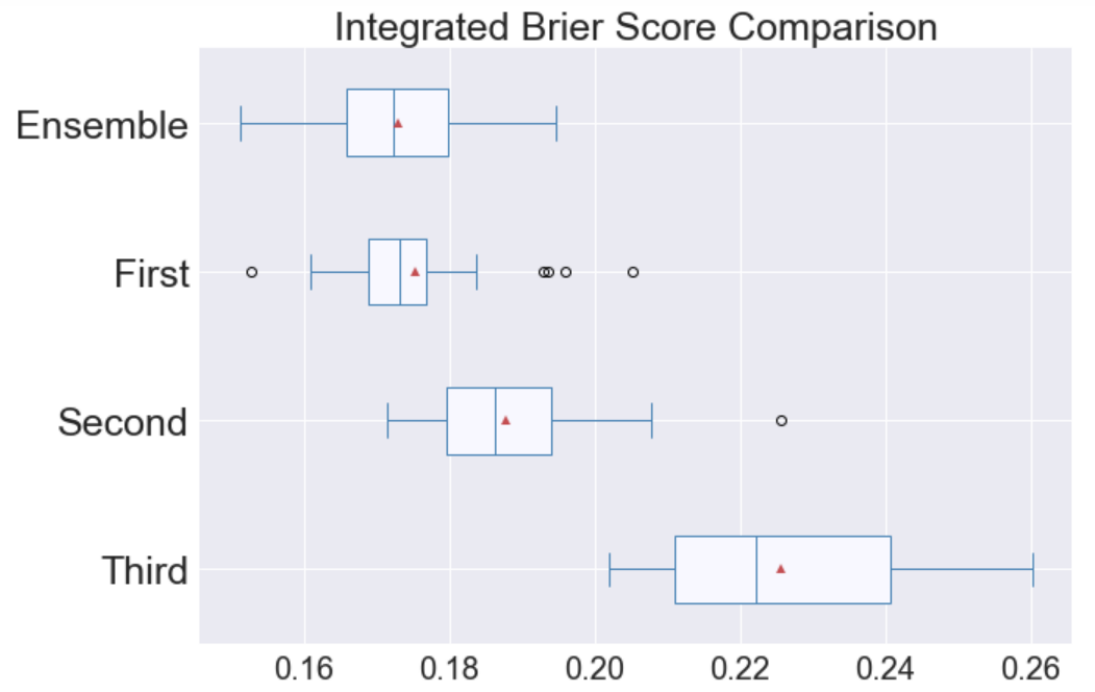}
 \caption{Integrated Brier score overall comparison among the three datasets} 
 \label{fig:test11}
\end{subfigure}
\caption{}
\end{figure}

\medskip
\noindent
Figure \ref{fig:test10} shows the comparison among all the techniques, including the deployed ensemble method. We see that, in the overall score, the ensemble method  outperforms the best predictor by $3.4 \%$. In Figure \ref{fig:test11}, we show the result obtained by averaging the best scores obtained by the six methods (described in Section \ref{sec3}) in each of the three datasets (they are DeepSurv for PBC and GBCSG2 and Cox PH for TLCM, see Figures \ref{fig:test4}, \ref{fig:test5} and \ref{fig:test6}, respectively) to obtain a global score, which corresponds to the average of the best scores among the six algorithms without using the ensemble method. We similarly  average the second best scores obtained by the six methods among the three datasets. Finally, this is also applied to the third best scores in the same way. The results are labeled as ``First'', ``Second'' and ``Third'' in Figure \ref{fig:test11}, respectively. We see that the ensemble method improves by $1\%$ the performance of the ``First'' score and has shown its effectiveness.

\section{Simulation Experiments}
\label{sec:simexp}

To deepen the insights from Section \ref{sec6}, we conducted experimental simulations with the goal of comparing the ranking of the methods under different dataset configurations; thus, to understand why some methods perform better than others. We simulate data using three different methods and under three different scenarios. The first two methods were based on R libraries coxed \cite{harden2019simulating} and simsurv \cite{brilleman2021simulating}. Both assume a particular shape of the hazard function, Cox proportional hazards and Weibull AFT respectively. The third method was created by us following the logic of the truck dataset from O. Grisel and V. Maladiere \cite{Grisel2023}. We give a further explanation in Section \ref{subsec:datasim}. This method was carried out in Python. The presented results are the averages obtained from 100 simulations. We do the comparison using the concordance index, and a similar analysis is presented in Appendix \ref{app:simexp} using the integrated brier score.

\medskip
\subsection{Python dataset simulation}
\label{subsec:datasim}
Following the truck dataset simulation from \cite{Grisel2023}, we first generate a specified number of features $d$ for each of the $N$ individuals. These features include normally distributed $\mathcal N (1,0.3)$ values, uniform $\mathcal U (0,1)$ values and categorical features of $3$ categories. Next, we define three types of failure, as mentioned in \cite{Grisel2023}: initial assembly failure, operation failure and fatigue failure. Although our method aims to be more general than the truck problem, we maintain the distributions specified in the cited reference. Each type of failure is modeled by a different Weibull curve with parameter $\lambda$. The first type of failure has a decreasing hazard with $\lambda = 0.003$, while the other two types have hazard rates that increase, with $\lambda = 3$ and $\lambda = 6$, respectively. The influence of the features on each of the failure types will vary in each experiment, depending on the number of features considered. To continue, we sample events of the three types for each individual and we choose the first one that occurs, or none if no event has taken place (censored case). Finally, we incorporate non-informative uniform censoring, where the parameters of the uniform distribution vary for each simulation case. The length of the uniform interval is what provides us with control over the percentage of censorship.


\medskip
\subsection{Number of samples}
\label{subsec:ns}

In this section, we compare the behavior of the methods as the number of samples increases. We consider $12$ features and $50\%$ of censorship. We vary the number of samples over a grid in between $50$ and $2000$ to study the impact of the number of samples in the performance of the different methods. The results are presented in the following figures.

\begin{figure}[H]
\centering
\begin{subfigure}[t]{0.49\textwidth}
  \includegraphics[width=\linewidth]{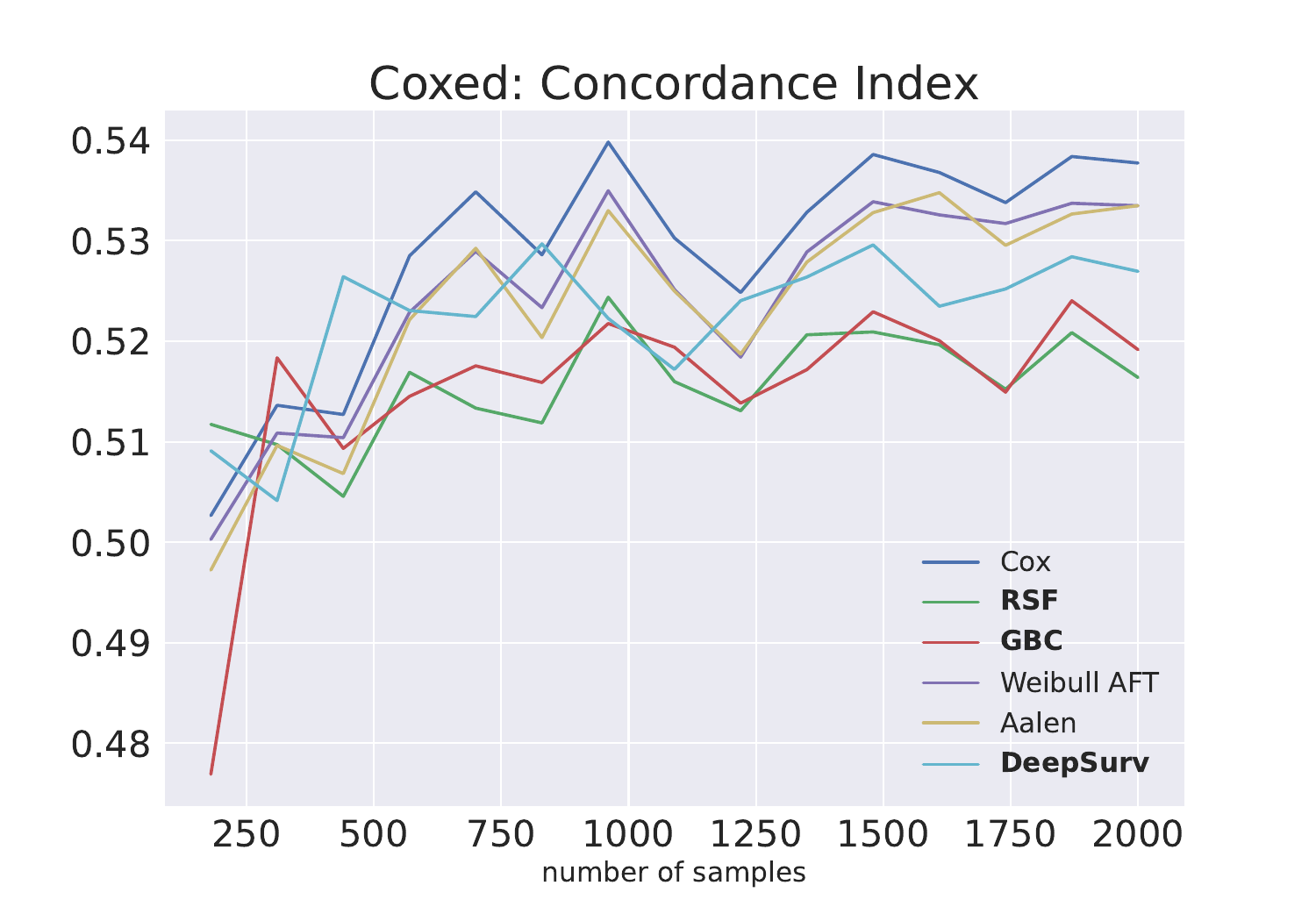}
  \caption{Concordance index comparison of the increasing sample size simulation with Coxed library}
  \label{fig:coxed_nsci}
\end{subfigure}
\hfill
\begin{subfigure}[t]{0.49\textwidth}
  \includegraphics[width=\linewidth]{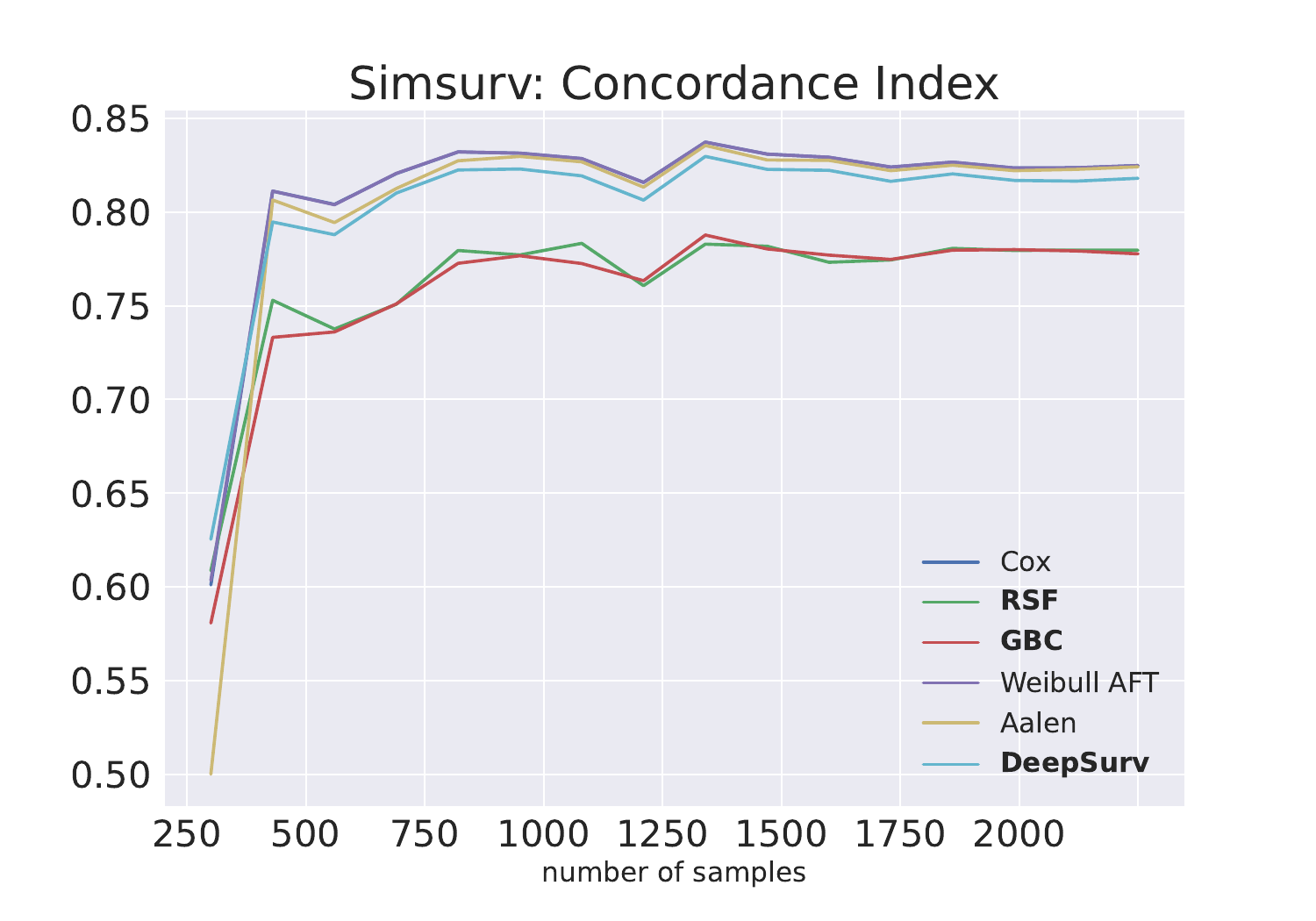}
  \caption{Concordance index comparison of the increasing sample size simulation with Simsurv library}
  \label{fig:simsurv_nsci}
\end{subfigure}
\caption{}
\end{figure}

\begin{figure}[H]
\centering
 \includegraphics[width = 0.49\linewidth]{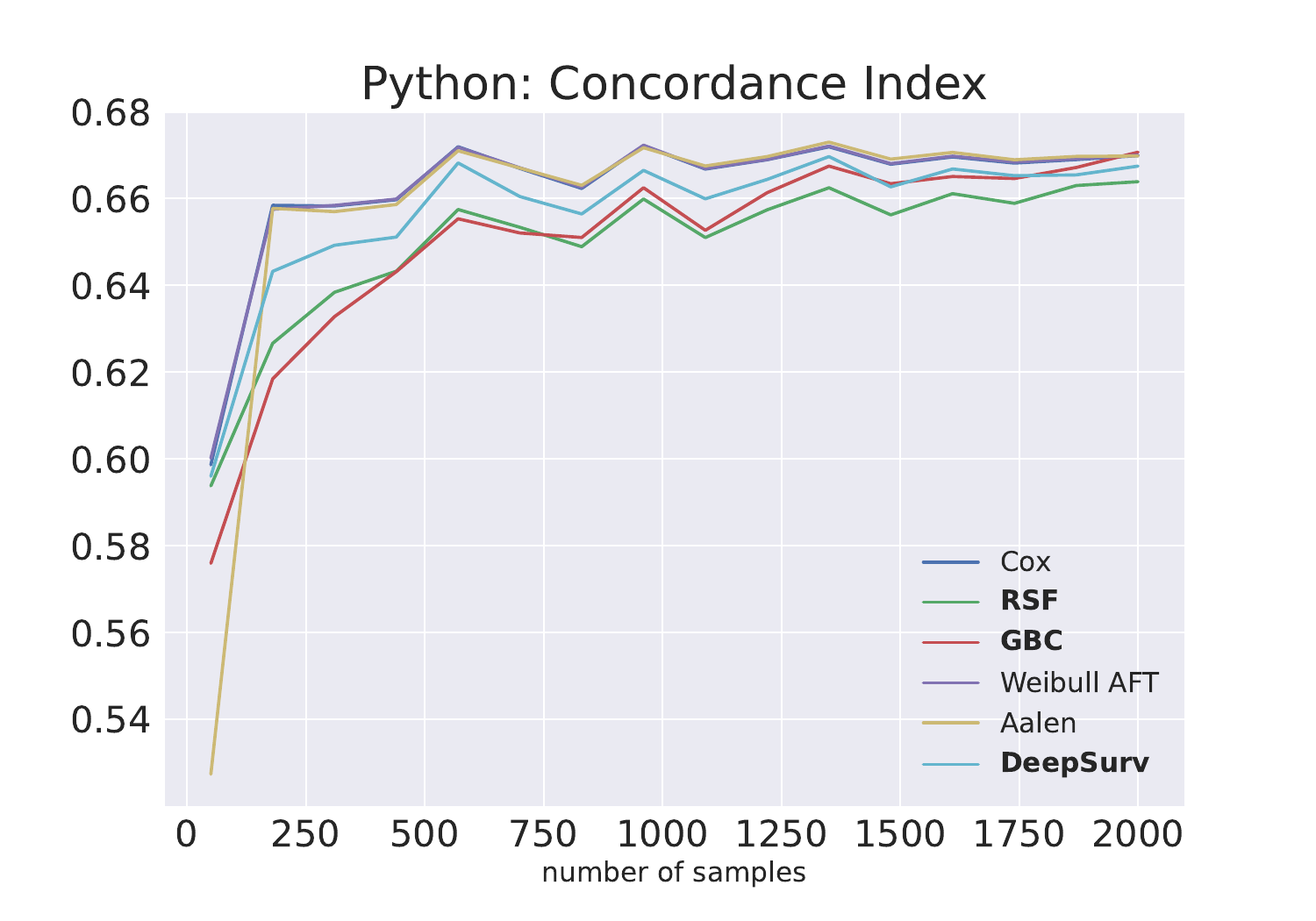}
\caption{Concordance index comparison of the increasing sample size simulation with Python}
  \label{fig:python_nsci}
\end{figure}

\noindent
We observe in Figures \ref{fig:coxed_nsci}, \ref{fig:simsurv_nsci}, and \ref{fig:python_nsci} that the concordance index improves as the number of samples increases. Additionally, we observe in Figure \ref{fig:coxed_nsci} that Cox proportional hazard consistently outperforms the other methods, regardless of the number of samples. Subsequently, the order is not very clear, but random survival forest and gradient boosting consistently show lower performance. In Figures \ref{fig:simsurv_nsci} and \ref{fig:python_nsci}, we observe a consistent outperformance of Cox proportional hazards and Weibull AFT, closely followed by Aalen additive hazards. Random
survival forest and gradient boosting underperform compared to the other methods in both figures. In conclusion, the ranking of the models performance appears to depend on the shape of the underlying distribution used to sample the event times and not on the number of samples of the dataset.

\subsection{Number of features}

In this section, we compare the behavior of the methods as the number of features decreases. With a fixed $50\%$ of censorship and $1000$ samples, we start the analysis with $20$ features and progressively remove one feature at each step. The results are presented in the following figures.

\begin{figure}[H]
\centering
\begin{subfigure}[t]{0.49\textwidth}
  \includegraphics[width=\linewidth]{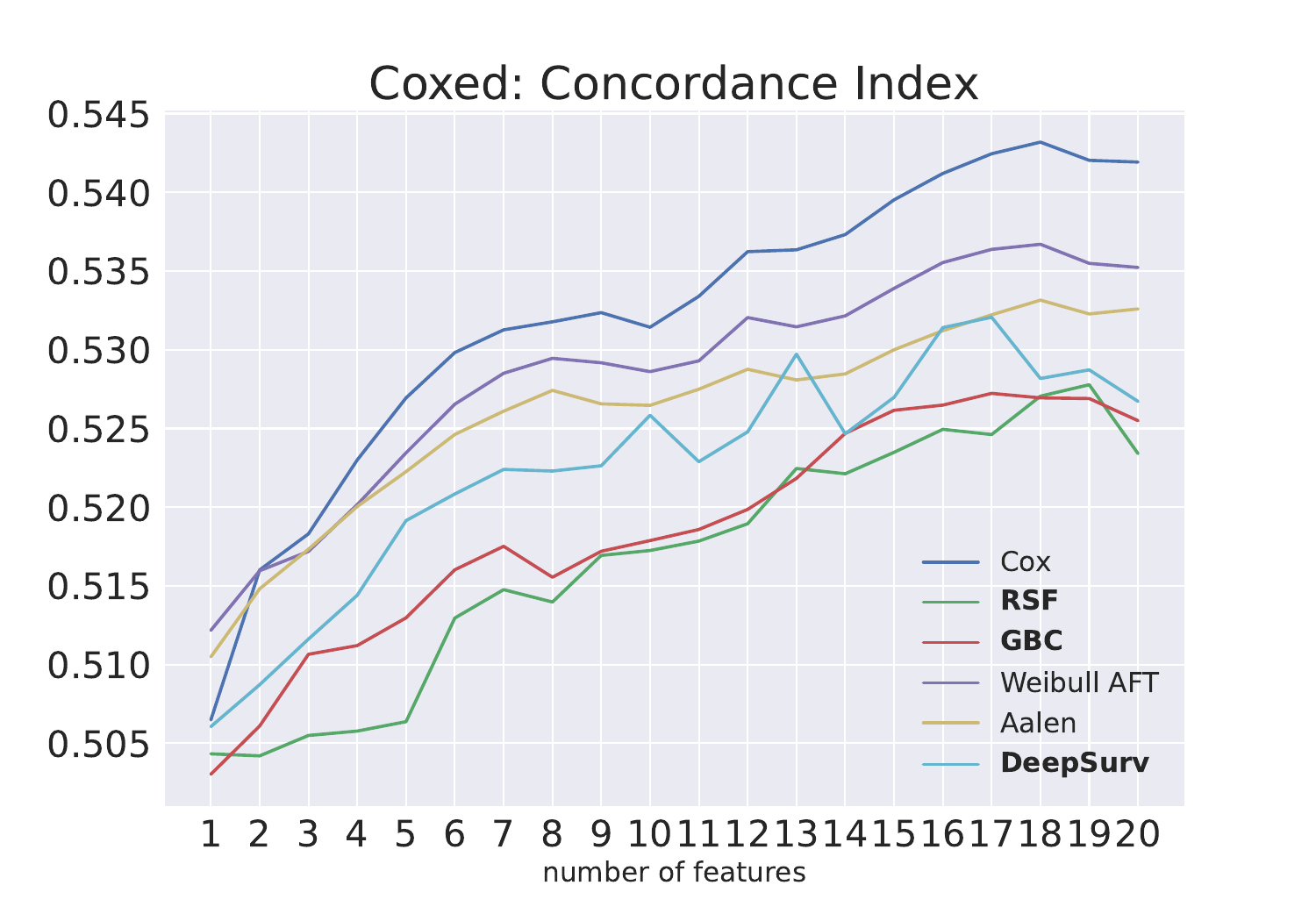}
  \caption{Concordance index comparison of the decreasing number of features simulation with Coxed library}
  \label{fig:coxed_nfci}
\end{subfigure}
\hfill
\begin{subfigure}[t]{0.49\textwidth}
  \includegraphics[width=\linewidth]{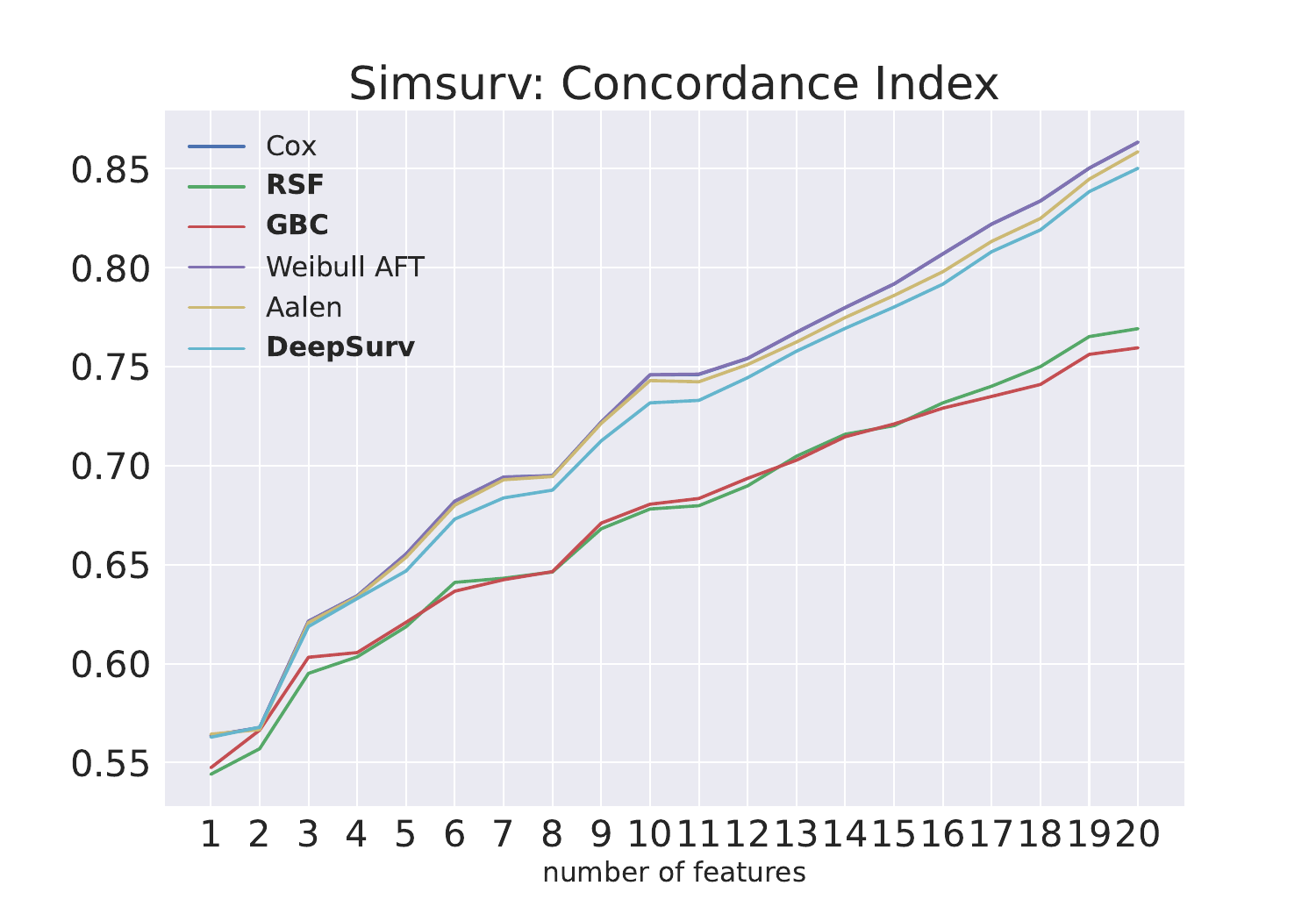}
  \caption{Concordance index comparison of the decreasing number of features simulation with Simsurv library}
  \label{fig:simsurv_nfci}
\end{subfigure}
\caption{}
\end{figure}

\begin{figure}[H]
\centering
 \includegraphics[width = 0.49\linewidth]{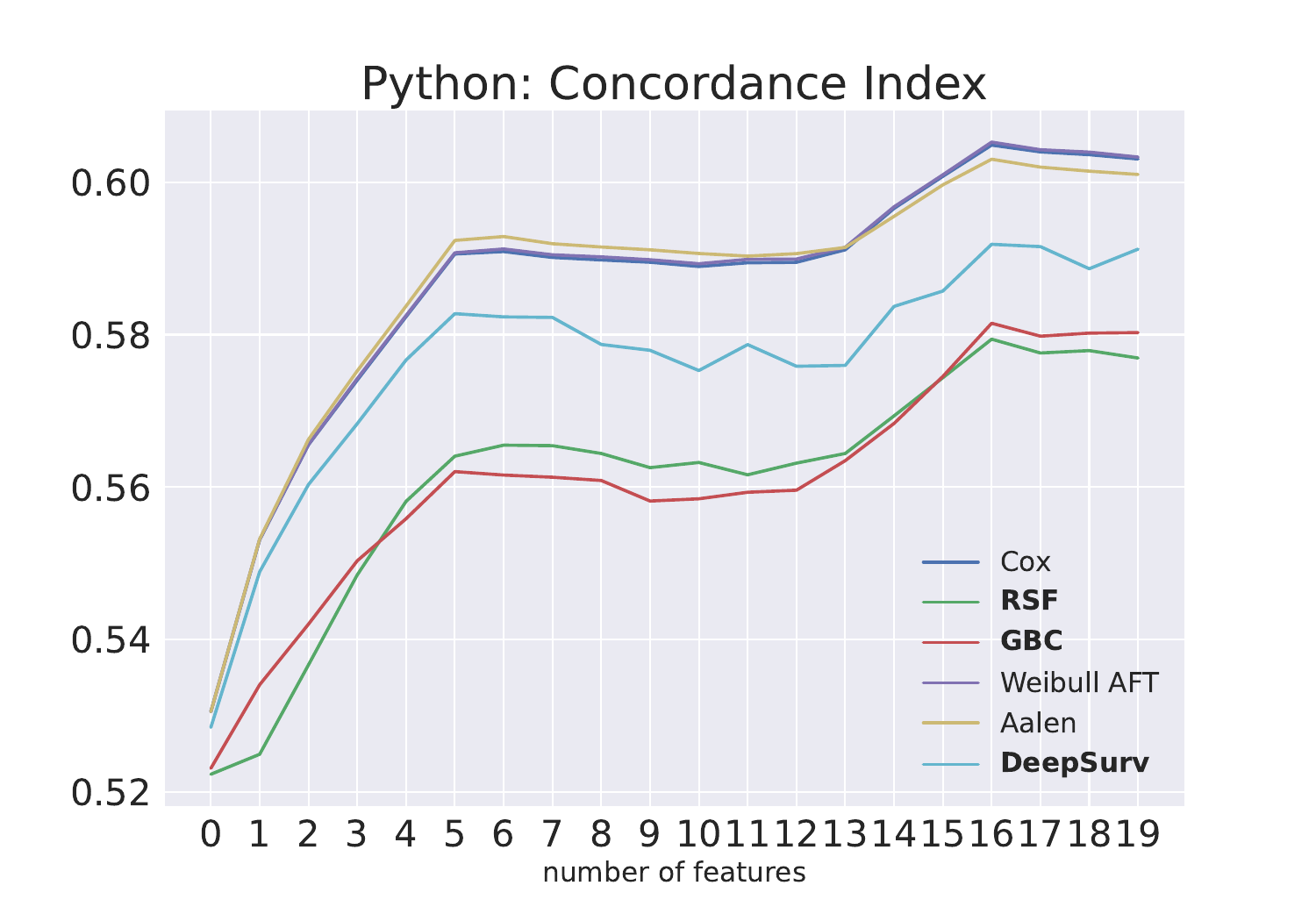}
  \caption{Concordance index comparison of the decreasing number of features simulation with Python}
  \label{fig:python_nfci}
\end{figure}

\noindent
We note in Figures \ref{fig:coxed_nfci}, \ref{fig:simsurv_nfci}, and \ref{fig:python_nfci} that the concordance index improves as the number of features increases. This behavior align with our expectations since the initial model is constructed with $20$ features, and the subsequent removal of features results in a reduction of information. Moreover, we observe in Figure \ref{fig:coxed_nfci} that, as in Figure \ref{fig:coxed_nsci}, Cox proportional hazard consistently outperforms the other methods, regardless of the number of features. The same holds  for Figure \ref{fig:simsurv_nfci} and \ref{fig:python_nfci}, where the best performance is shared by Cox proportional hazard, Weibull AFT, and Aalen additive hazards. Following the conclusion of Section \ref{subsec:ns}, the ranking of the models depends mainly on the shape of the distribution used to generate the data, rather than on the number of features.

\subsection{Percentage of censorship}

In this section, we compare the behavior of the methods as the percentage of censorship increases. We fix the number of samples at $1000$ and the number of features at $12$. The results are presented in the following figures.

\begin{figure}[H]
\centering
\begin{subfigure}[t]{0.49\textwidth}
  \includegraphics[width=\linewidth]{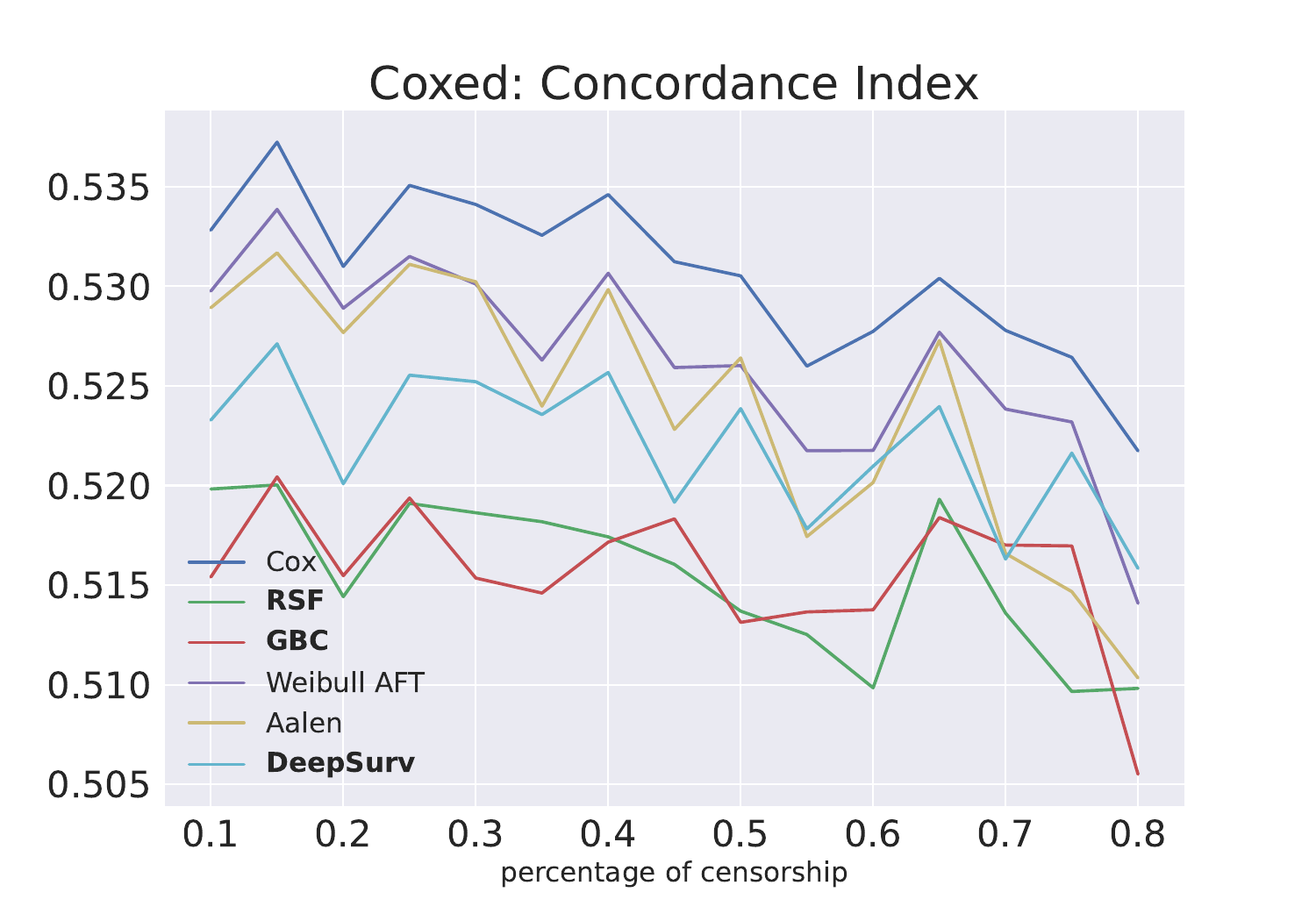}
  \caption{Concordance index comparison of the increasing percentage of censorship simulation with Coxed library}
  \label{fig:coxed_pcci}
\end{subfigure}
\hfill
\begin{subfigure}[t]{0.49\textwidth}
  \includegraphics[width=\linewidth]{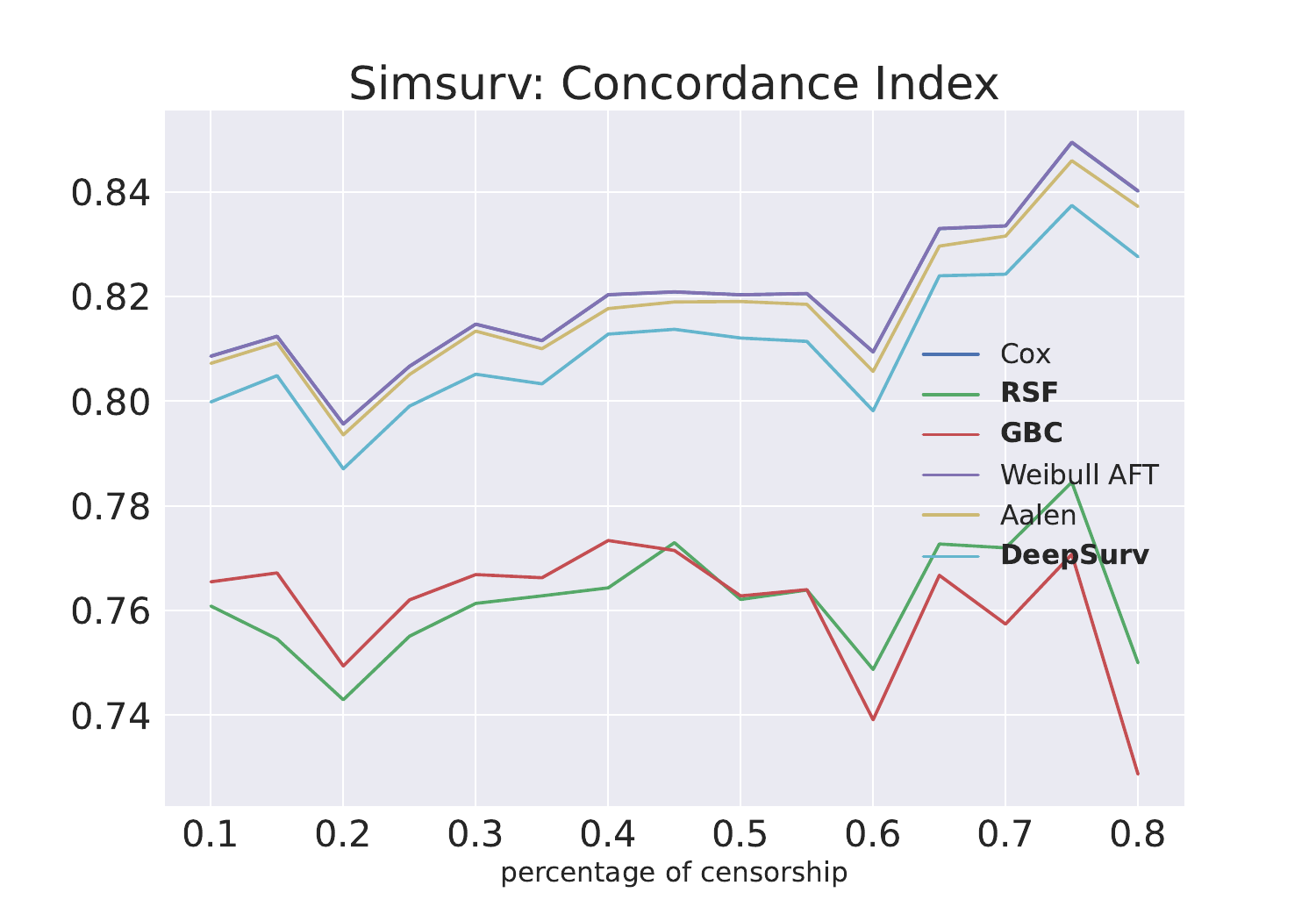}
  \caption{Concordance index comparison of the increasing percentage of censorship simulation with Simsurv library}
  \label{fig:simsurv_pcci}
\end{subfigure}
\caption{}
\end{figure}

\begin{figure}[H]
\centering
 \includegraphics[width = 0.49\linewidth]{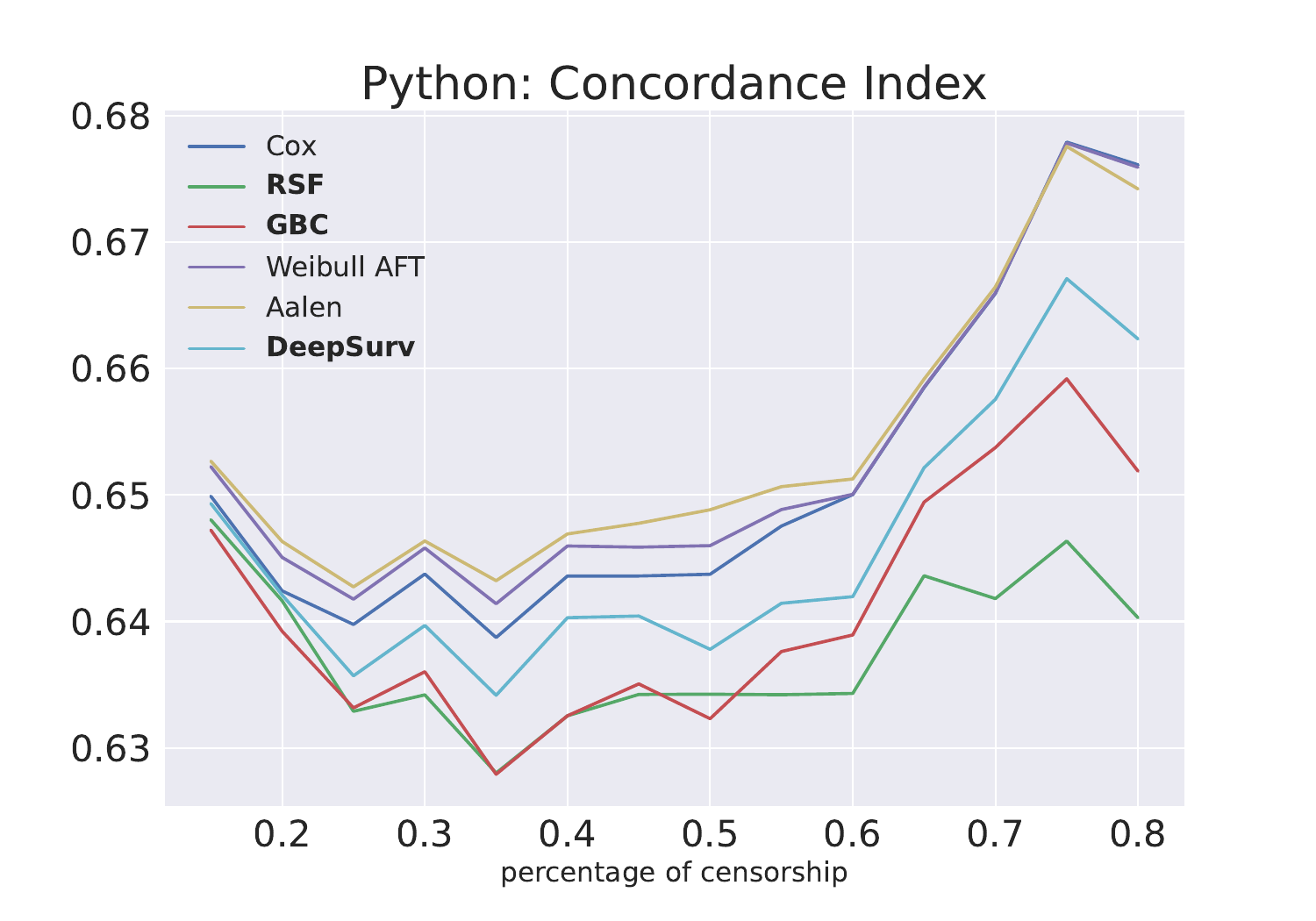}
  \caption{Concordance index comparison of the increasing percentage of censorship simulation with Python}
  \label{fig:python_pcci}
\end{figure}

\noindent
In Figure \ref{fig:coxed_pcci}, we observe a decline in performance as the percentage of censorship increases. This is in line with the notion that higher levels of censorship result in reduced available information, consequently leading to diminished performance. However, a contrasting pattern emerges in Figure \ref{fig:simsurv_pcci} and \ref{fig:python_pcci}, where we actually observe an improvement in performance towards the end of the curves. We believe that this phenomenon is attributed to a bias in the concordance index when the percentage of censorship is high. One solution to address this issue is presented by Uno et al. \cite{uno2011c}, where they introduced a weighted version of the score. In addition, we observe in Figure \ref{fig:coxed_pcci} that Cox proportional hazards outperforms the other methods, followed by Weibull AFT and Aalen additive. This same pattern is evident in Figures \ref{fig:simsurv_pcci} and \ref{fig:python_pcci}, where these three models lead in terms of performance. Notably, the ranking of the methods remains consistent even as the percentage of censorship increases, reinforcing the conclusion from the previous sections. The primary factor influencing the performance change of the methods is the congruence between the model assumptions and the actual distribution of event times, with improved fit leading to better performance.\\

\section{Conclusions}
\label{sec7}
This paper presents an extensive analysis of different survival methods applied to three datasets and compared by two scoring rules. The study shows how diverse a single method's performance is when changing the measure of comparison and when it is applied to datasets of different distributions, sizes and percentages of censorship. We propose a straightforward aggregation of methods of different natures, parametric, semi-parametric and machine learning, that assume diverse shapes of the hazard function allowing the ensemble model to gain in robustness with respect to each single predictor. This can be observed in Figure \ref{fig:test10} by the outperformance of the assemblage measured by an overall score that is independent of the dataset. Finally, we present simulation experiments with the objective of studying which dataset characteristics have the most significant influence on the performance of the models. This analysis leads us to the conclusion that the proximity of the model assumptions to the real event distribution is a determining factor in performance. Further research could go in the direction of complexifying the combination algorithm by considering time-varying weightings and more sophisticated optimization procedures. Another direction could be to find theoretical guarantees for the integrated Brier score of the ensemble method and possibly in a stochastic setting.\\ 


\bibliography{bibliography}

\bigskip
\appendix

\section{Scoring Rules}
\label{App1}

\subsection{Concordance Index}
\label{subsec21}

The concordance index was introduced by Harrell et al. \cite{harrell1996multivariable} and it is the most widely used performance metric for time-to-event analysis \cite{steck2008ranking}. It measures the fraction of pairs of subjects that are correctly ordered within all the possible pairs that can be ordered. The highest (and best) value that can be obtained is $1$, which means that there is a complete agreement between the order of the observed and predicted times. The lowest value that can be obtained is $0$, which means that all the prediction pairs are ordered backward with respect to the observed times, while a value of $0.5$ denotes a random model. 

\medskip
\noindent
First, we take every pair in the test set such that the earlier observed time is not censored. Then, we consider only pairs $(i,j)$ such that $i<j$ and we also eliminate the pairs for which the times are tied. Next, we define a score $C_{i,j}$ for each pair $(i,j)$ such as $y_i \neq y_j$, equal to $1$ if the subject with earlier time (between $i$ and $j$) has higher predicted risk, equal to $0.5$ if the risks are tied, or equal to $0$ otherwise.

\medskip
\noindent
Finally, given a subset of the data $\mathcal{D}$ of size $n$, we compute the concordance index as follows: 
\begin{equation} 
CI(\hat{S}, \mathcal{D}) =  \frac{1}{\vert \mathcal{P}\vert} \sum\limits_{(i,j) \in \mathcal{P}} C_{i,j}, \nonumber
\end{equation}
where,
\begin{align} 
C_{i,j} =  \begin{cases} 
      1 & \text{if } y_i < y_j \text{ and } \hat{R}(x_i) >\hat{R}(x_j)   \\
      0.5 & \text{if } \hat{R}(x_i) = \hat{R}(x_j)   \\
      0 &  
      \text{otherwise,} 
   \end{cases} \nonumber
\end{align}
and $\mathcal{P} = \{ (i,j) \in \mathcal{D} \times \mathcal{D} : i < j, ~ y_i \neq y_j, ~ \text{if} ~ y_i < y_j \text{, then} ~ \delta_i = 1  \}$ is the set of all eligible pairs. To calculate the concordance index, we use the version of scikit-survival library \cite{polsterl2020scikit} in Python.

\medskip
\subsection{Integrated Brier score}
\label{subsec22}

We consider an approach based on the estimates of the probability functions that will be used as predictions of the event status $\mathds{1}\{T_i > t\}$. In this case, $\mathds{1}\{T_i > t\}$ has to be compared with $\hat{S}(t\vert X_i)$, leading to the mean squared error ($MSE$) at time $t$:
\begin{equation} 
MSE(\hat{S},t) = \mathbb{E}[(\mathds{1}\{T_i>t\} - \hat{S}(t\vert X_i) )^2]. \nonumber 
\end{equation}

\noindent
The Brier score, introduced initially to measure the inaccuracy of probabilistic weather forecast by Brier \cite{brier1950verification}, is an estimator of the $MSE$. It is important to remark that the $MSE$ cannot be directly computed from the dataset since we do not know the underlying distribution of $T_i$ but only the realizations of $Y_i$. Let us define $S_C(t\vert X_i) = \mathbb{P}(C_i>t\vert X_i)$ the survival censoring distribution and the Brier score:
\begin{equation}
BS(\hat{S},t,\mathcal{D}) = \frac{1}{n} \sum\limits_{i=1}^{n}  W_i(t) (\mathds{1}\{ y_i > t) - \hat{S}(t\vert x_i))^2, \nonumber 
\end{equation}
where $(x_i,y_i, \delta_i)$ for $0<i\leq n$ are points from $\mathcal{D}$, and $W_i$ is defined for all $t$ as:
\begin{equation} 
W_i(t) = \frac{\delta_i \mathds{1}\{y_i \leq t\} }{\hat{S}_C(y_i\vert x_i)} + \frac{\mathds{1}\{y_i > t\}}{\hat{S}_C(t\vert x_i)}. \nonumber
\end{equation}

\noindent
Gerds and Schumacher \cite{gerds2006consistent} proved that the Brier score is a consistent estimator for the mean square error when the estimation $\hat{S}_C$ of $S_C$ is well specified. Let us notice that in our implementation of the score, we use a Kaplan-Meier \cite{kaplan1958nonparametric} estimator for the survival censoring function $\hat{S}_C$, which does not depend on the covariates. This assumption is not always the case for the real censoring function $S_C$, and it can lead to misspecifications of the model (wrong hypothesis on the probability shape) and, thus, to an estimation bias \cite{graf1999assessment}.

\medskip
\noindent
Finally, we consider over this paper the integrated Brier score: \begin{equation}
IBS(\hat{S},\mathcal{D}) = \frac{1}{\tau} \int\limits_{0}^{\tau} BS(\hat{S},t,\mathcal{D}) dt, \nonumber 
\end{equation}
where $\tau$ is a user-specified time horizon. There exist diverse scoring rules for survival models based on $L1$-loss, logarithmic loss and 1-calibration in between others (see \cite{haider2020effective} and \cite{graf1999assessment} for more details). Other approaches of the estimation of prediction errors and model misspecification can be found in \cite{lawless2010estimation}. We chose the concordance index and integrated Brier score because they measure different aspects of the models, ranking and calibration, allowing us to have a good understanding of the performance of the methods.\\

\section{Implemented Methods}
\label{App2}
 
\subsection{Cox Proportional Hazard (Cox PH)}
\label{subsec31}

Cox proportional hazard is a semi-parametric method proposed by Cox \cite{cox1972regression} with the objective of measuring the impact of each covariate/feature in the estimation of the survival probability function. It models the hazard function as a general linear regression of the covariates and a non-parametric baseline function $\lambda_0(t)$ that depends only on time. Given a subject with a covariate vector $x = \{x^1, \ldots, x^d \}$, the hazard function is as follows:
\begin{equation} 
h(t\vert x) = \lambda_0(t) \exp \left(   \beta^T x  \right), \nonumber
\end{equation}
where the parameter $\beta = (\beta_1,...,\beta_d)$ is estimated by maximizing the likelihood. This model is semi-parametric in the sense that the baseline function $\lambda_0(t)$ does not need to be specified and it can be chosen differently for each unique time. Cox proportional hazard is one of the most often used methods in time-to-event analysis and has a wide range of applications \cite{lane1986application}, \cite{liang1990cox}, \cite{sauerbrei1992bootstrap}. We use the implementation from scikit-survival library \cite{polsterl2020scikit}, where a regularization parameter $\alpha$ for ridge regression penalty is used and whose default is equal to $0$. The mortality risk prediction will be determined by the log hazard ratio $R(x) = \beta^Tx$. 

\medskip
\subsection{Gradient Boosting Cox (GBC)}
\label{subsec32}

Gradient boosting Cox is a machine learning method that was first proposed by Breiman \cite{breiman1997pasting}, developed by Friedman \cite{friedman2001greedy} and adapted to survival models by Ridgeway \cite{ridgeway1999state}. The main idea is to combine a series of base learners in an additive manner to obtain a strong overall model. The base learners implemented in this case will be regression trees fitted at each stage on the negative gradient of the loss function. This is an additive method in the sense that it is constructed sequentially in a step-by-step greedy way. We can define the overall function $f$ as follows:
\begin{equation} 
f(x) = \sum\limits_{k=1}^K \rho_k \cdot g_k(x,\theta), \nonumber
\end{equation}
where $g_k$ is used to denote the base learners and $K$ is the number of learners. Therefore, the objective is to maximize the log-likelihood function of Cox's proportional hazard model by replacing the linear regression $\beta^T x$ with the additive function $f(x)$ such that we have the following expression for the hazard function: 
\begin{equation}
h(t\vert x) = \lambda_0(t) \exp \left(f(x)\right). \nonumber
\end{equation}
We use the implementation of scikit-survival \cite{polsterl2020scikit} where we find three parameters of our interest, the learning rate that shrinks the contribution of each tree and it is set as default by $0.1$, the maximum depth that specifies the depth to which each tree will be built and that is set equal to $3$ by default, and the minimum samples leaf that determines the number of samples required to be at a leaf node and its default is equal to $1$. Similar as Cox proportional hazard the mortality risk prediction can be interpreted as the log hazard ratio $f(x)$.

\medskip
\subsection{Random Survival Forest (RSF)}
\label{subsec33}

Random survival forest was proposed by Ishwaran et al. \cite{ishwaran2008random} as an adaptation for censored data of the random forest method introduced by Breiman et al. \cite{breiman2001random}. It is an ensemble of tree-based learners where each tree is built from a bootstrap sampling of the training set in order to reduce the correlation between the trees. Also, for each node, it only evaluates the split criterion for a random subset of features and thresholds. The quality of a split is measured by the log-rank splitting rule \cite{bland2004logrank} and then predictions are formed by aggregating predictions of the individual trees.

\medskip
\noindent
We implemented random survival forest from scikit-survival \cite{polsterl2020scikit} and we will consider three of its parameters. The first one is the maximum depth which is set as infinity, which means that the nodes are expanded until no further partitioning is possible. The second one is the maximum features number which indicates the maximum number of features to consider when looking for the best split; this parameter is set as the number of data features. The last parameter is the minimum samples leaf which in this case the default value is $3$. Here, the mortality risk is defined by the ensemble mortality (see \cite{ishwaran2008random} for details) which corresponds to the sum of the cumulative Hazard functions estimated by the forest.  

\medskip
\subsection{Weibull Accelerated Failure Time (Weibull AFT)}
\label{subsec34}

Weibull AFT is a parametric model that was named after Waloddi Weibull, who was the first to promote its usefulness, particularly in the domain of strength of materials \cite{weibull1939statistical}. Accelerated failure time models also assume that the effect of a covariate is to accelerate or decelerate the life course. Given the parameters $\rho$ and $\lambda$, the survival function of the Weibull distribution 
can be given as:
\begin{equation}
S(t\vert x) = \exp \left( - \left( \frac{t}{\rho(x)} \right)^{\lambda} \right), \nonumber
\end{equation}
where we consider the scale parameter  $\rho (x) = \exp \big(\beta_0 \cdot ( \beta^T x )  \big) $  and $\lambda$ is the parameter that controls the concavity of the cumulative hazard, indicating acceleration or deceleration hazards. In this case, we implement Weibull AFT from lifelines library \cite{davidson2019lifelines} and we consider a penalizer parameter and a $\ell1$-ratio to adjust how much of the penalizer should be attributed to an $\ell1$ penalty. Both of them are initially set as zero by default. Here, the mortality risk is defined by $\mathbb{E}[T_i\vert x_i]$.

\medskip
\subsection{Aalen's Additive Fitter (Aalen)}
\label{subsec35}

Aalen's additive is a parametric method proposed by Aalen \cite{aalen1989linear}. This model responds to the fact that not all the covariates effects must be proportional, which is different from the assumption of Cox proportional hazard, but some of them can have additive effects. Besides, Aalen's additive model allows the effects of the covariates to vary over time which is not always the case with the other methods. The hazard function, in this case, is given as follows:
\begin{equation}
h(t\vert x)= \beta_0 (t) +  \beta^T(t) x, \nonumber
\end{equation}
where $\beta(t)$ is an unknown parameter of dimension $d$ that are estimated by a linear regression (see \cite{aalen2005aalen}). We consider only the penalizer coefficient, which attaches an $\ell2$ penalizer to the size of the parameters during regression that improves the stability of the estimations and controls the high correlation of the features. This penalizer is set to zero by default. Similarly as Weibull AFT, the mortality risk is defined by $\mathbb{E}[T_i\vert x_i]$.

\medskip
\subsection{DeepSurv}
\label{subsec36}

DeepSurv is a nonlinear version of the Cox proportional hazard method proposed by Katzman et al. \cite{katzman2018deepsurv}. DeepSurv allows the use of neural networks within the original design of Cox's and aims to offer more flexibility in terms of the structure of the model than Cox proportional hazard. DeepSurv is a multi-layer perceptron that predicts the risk of failure. The output of the network $\hat{r}_{\theta}(x)$ is a single node that estimates the risk function. The loss function to minimize is the negative log-partial likelihood of the Cox proportional hazard method: 
\begin{align}
\ell(\theta) = & - \frac{1}{L} \sum\limits_{i:\delta_i=1} \left( \hat{r}_{\theta}(x_i)- \log \left( \sum\limits_{j \in \mathcal{R}(T_i)} e^{\hat{r}_{\theta}(x_j)} \right) \right) \nonumber\\ 
& + \lambda || \theta||^2, \nonumber
\end{align}
where $\lambda$ is a $\ell_2$ regularization parameter and $L$ is the number of uncensored subjects. The network weights that minimize the loss function can be estimated by a gradient descent algorithm~\cite{ruder2016overview}. We use the implementation from Pysurvival \cite{pysurvival_cite}, where we can choose the structure of the multilayer perceptron by choosing the number of hidden units per layer. We will consider two fully connected hidden layers and, consequently, two parameters to be set. The default number of units is 60 for the first layer and 10 for the second.\\

\section{Simulation Experiments}
\label{app:simexp}

In this section, we simulate data using three different techniques. The first set of events is generated by sampling a Cox proportional hazards model, the second by following a Weibull distribution, and the third involves a combination of Weibull distributions. The objective is to compare how the ranking of the methods varies across three experiments. Thus, to understand how different data characteristics can impact the performance of the methods. These findings align with those presented in Section \ref{sec:simexp}, with the distinction that we assess performance using the integrated Brier score.

\medskip
\subsection{Number of samples}
The first experiment consists on evaluating the performance of the models as the number of samples increases from $50$ to $2000$.

\begin{figure}[H]
\centering
\begin{subfigure}[t]{0.49\textwidth}
  \includegraphics[width=\linewidth]{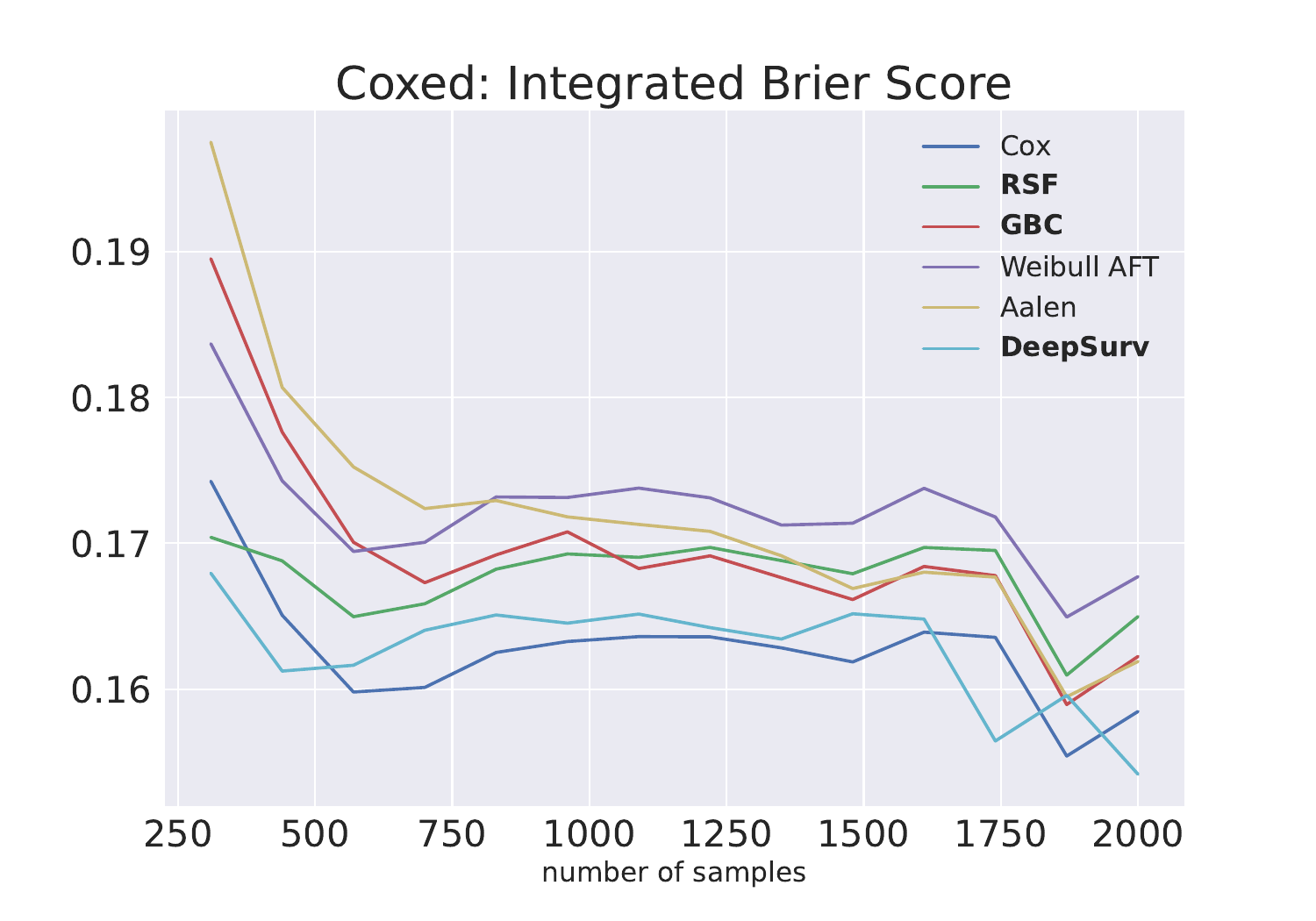}
  \caption{Integrated Brier score comparison of the increasing sample size simulation with Coxed library}
  \label{fig:coxed_nsibs}
\end{subfigure}
\hfill
\begin{subfigure}[t]{0.49\textwidth}
  \includegraphics[width=\linewidth]{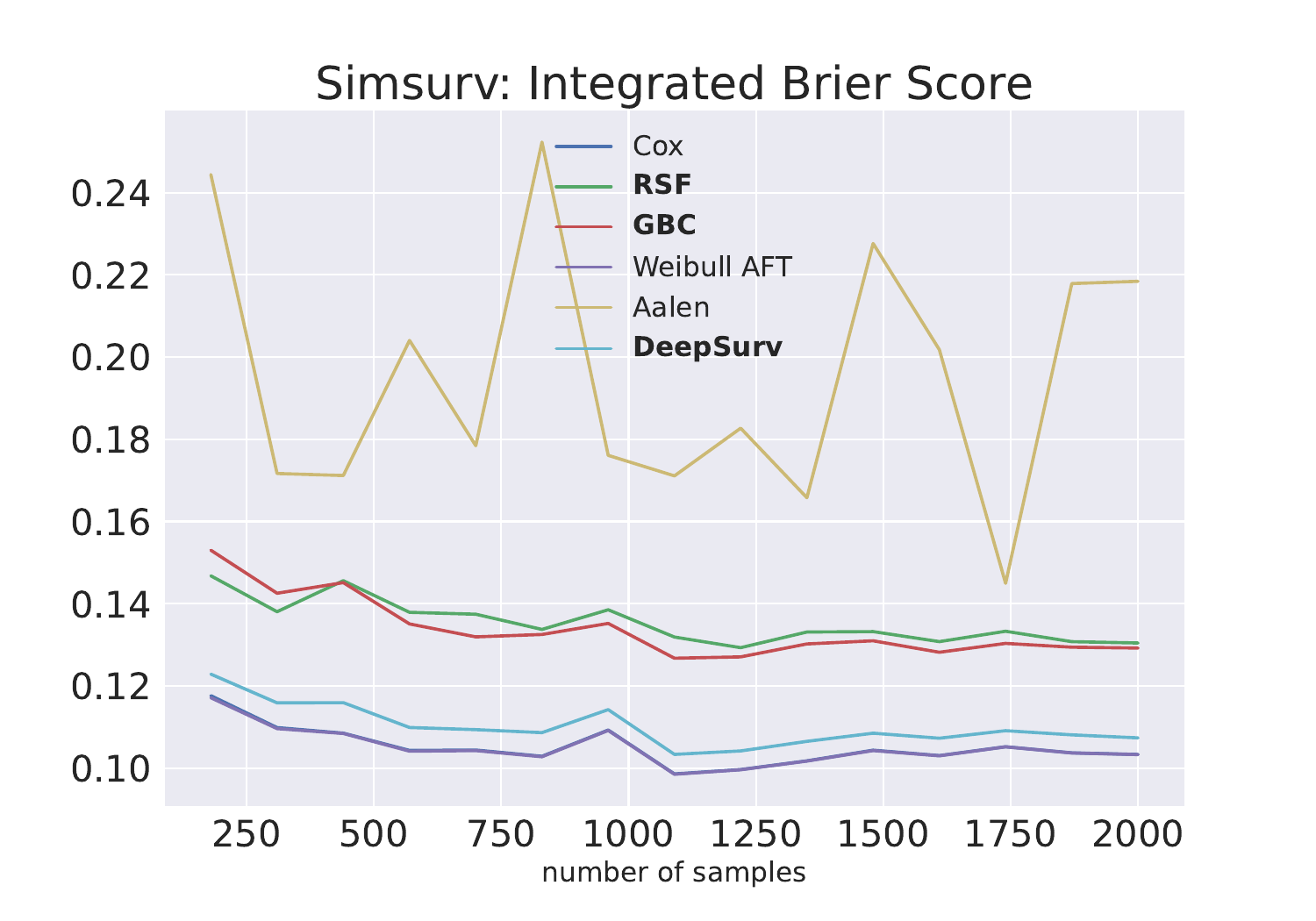}
  \caption{Integrated Brier score comparison of the increasing sample size simulation with Simsurv library}
  \label{fig:simsurv_nsibs}
\end{subfigure}
\caption{}
\end{figure}

\begin{figure}[H]
\centering
 \includegraphics[width = 0.49\linewidth]{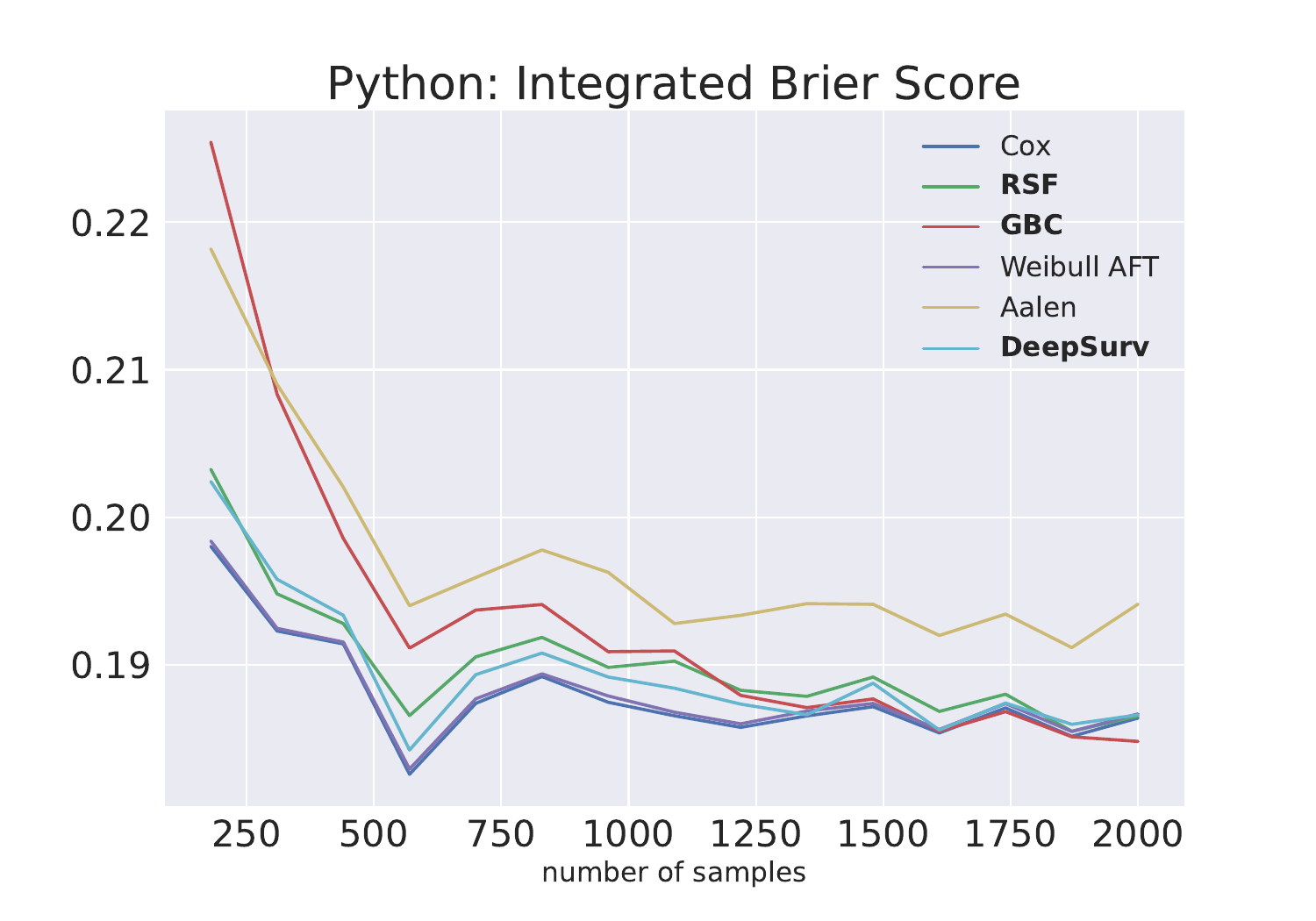}
  \caption{Integrated Brier score comparison of the increasing sample size simulation with Python}
  \label{fig:python_nsibs}
\end{figure}

We observe in Figures \ref{fig:coxed_nsibs}, \ref{fig:simsurv_nsibs}, and \ref{fig:python_nsibs}, as discussed in Section \ref{subsec:ns}, that the performance improves as the number of samples increases. Furthermore, it is noteworthy that the hierarchy of the models remains relatively stable as the number of samples increases. Specifically, Cox proportional hazards outperforms the other methods, with DeepSurv as the second-best performer. This reaffirms the conclusion made in Section \ref{subsec:ns} that the models' performance order is independent of the sample size but instead depends on the matching between the underlying assumptions and the dataset real distribution shape.

\subsection{Number of features}
The second experiment consists on evaluating the performance of the models as the number of features decreases from $20$ to $1$.

\begin{figure}[H]
\centering
\begin{subfigure}[t]{0.49\textwidth}
  \includegraphics[width=\linewidth]{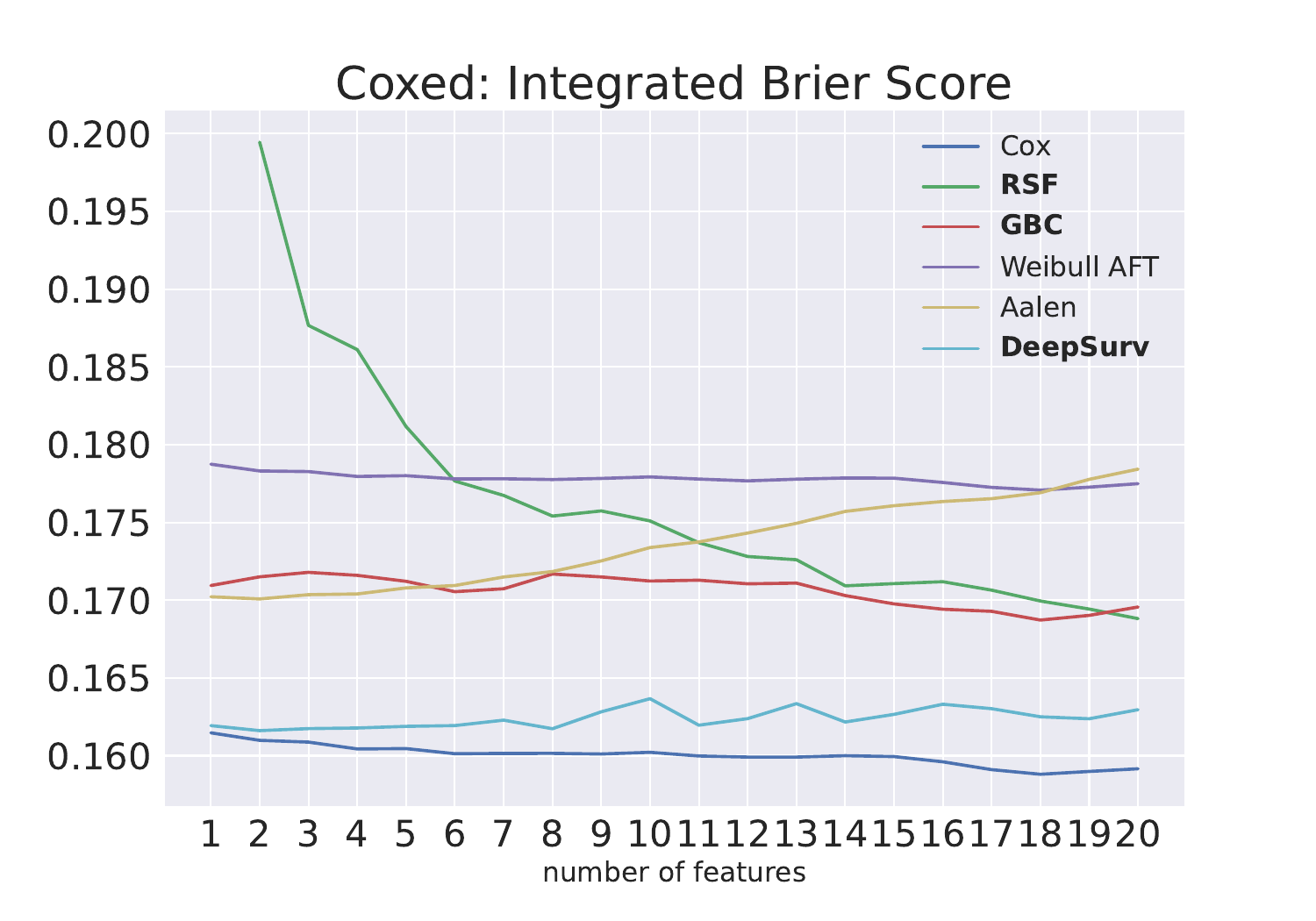}
  \caption{Integrated Brier score comparison of the decreasing number of features simulation with Coxed library}
  \label{fig:coxed_nfibs}
\end{subfigure}
\hfill
\begin{subfigure}[t]{0.49\textwidth}
  \includegraphics[width=\linewidth]{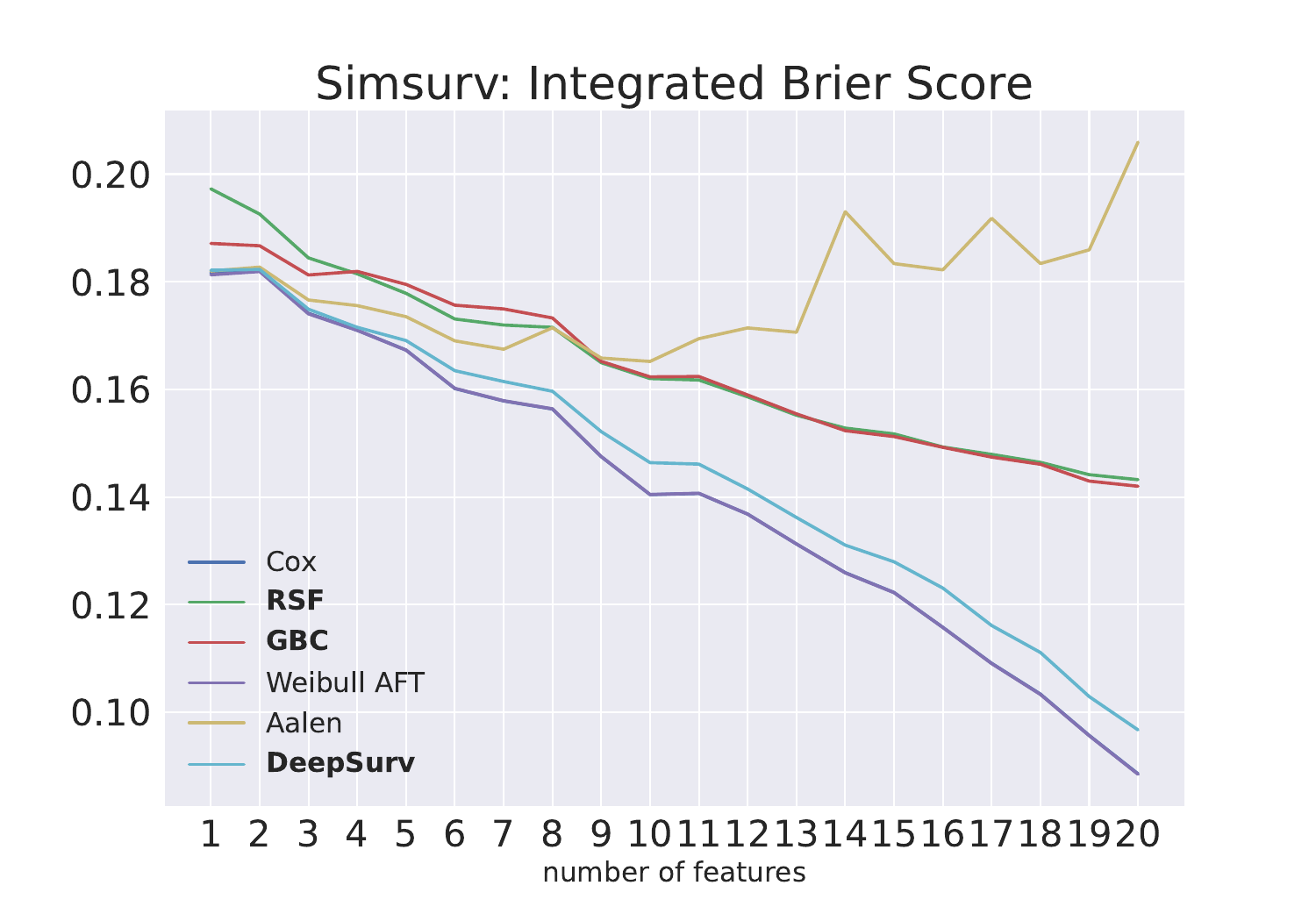}
  \caption{Integrated Brier score comparison of the decreasing number of features simulation with Simsurv library}
  \label{fig:simsurv_nfibs}
\end{subfigure}
\caption{}
\end{figure}

\begin{figure}[H]
\centering
 \includegraphics[width = 0.49\linewidth]{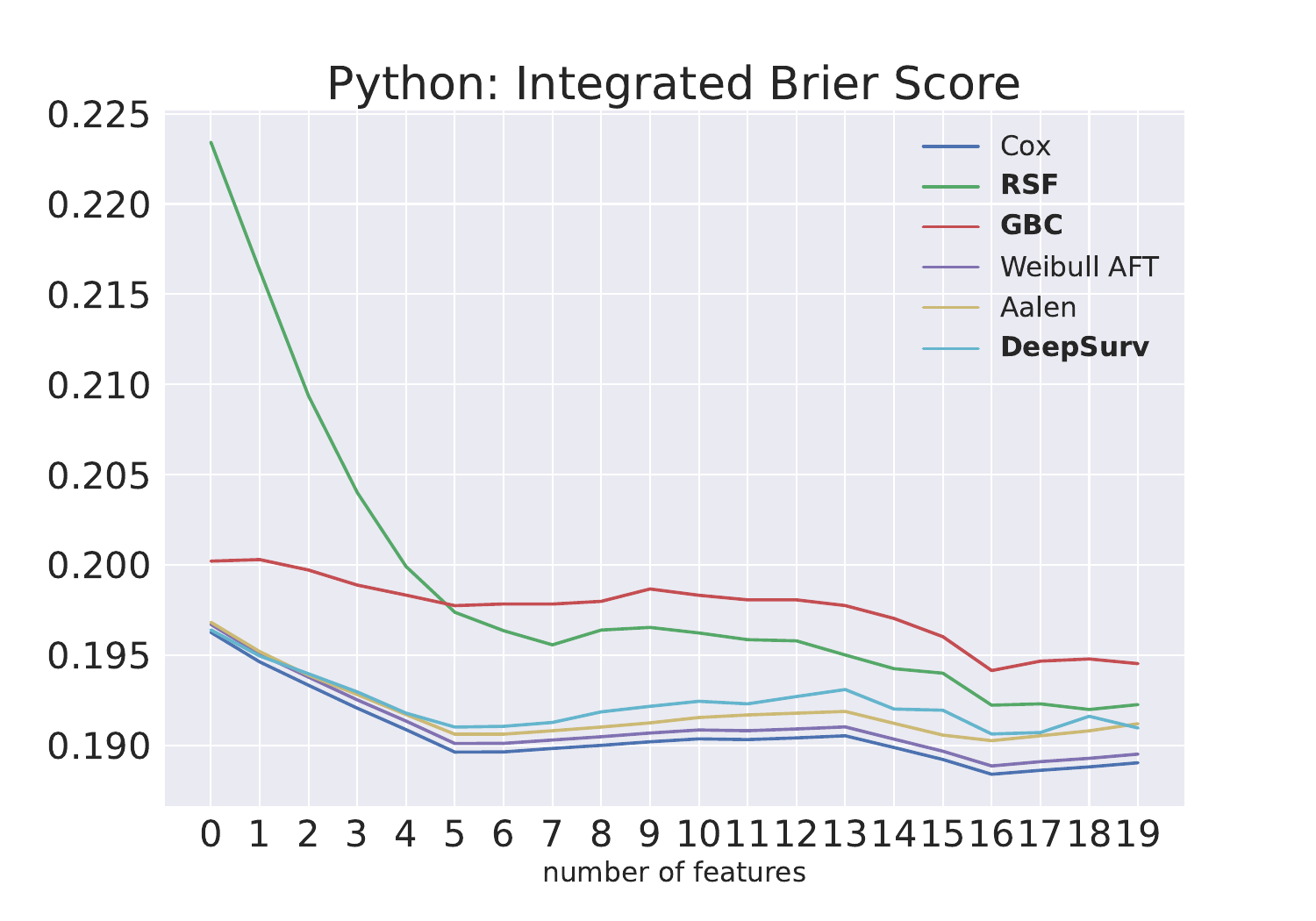}
  \caption{Integrated Brier score comparison of the decreasing number of features simulation with Python}
  \label{fig:python_nfibs}
\end{figure}

\noindent
We observe in Figures \ref{fig:coxed_nfibs}, \ref{fig:simsurv_nfibs} and \ref{fig:python_nfibs} a slight increase in performance as the number of features increases. The most significant change occurs with Random Survival Forest, which exhibits very poor performance compared to the other methods when the number of features is small but becomes competitive as the number of features increases. This is because random survival forest relies on the diversity and richness of features to make accurate predictions. In addition, in Figures \ref{fig:coxed_nfibs} and \ref{fig:simsurv_nfibs}, we observe that the Aalen additive model does not align with the trend of the other methods, as its performance worsens with an increasing number of features. This could be due to the challenge posed by the additive linearity assumption in capturing the true underlying relationship between covariates and survival outcomes. Finally, we observe that the performance of the Weibull AFT model improves relatively in Figures \ref{fig:simsurv_nfibs} and \ref{fig:python_nfibs} compared to Figure \ref{fig:coxed_nfibs}. This phenomenon occurs because both simulations, those conducted by the simsurv library and our method implemented in python, are based on the Weibull distribution.

\subsection{Percentage of censorship}
The third experiment, as presented in Section \ref{sec:simexp}, consists on increasing the percentage of censorship from $10\%$ to $80 \%$.

\begin{figure}[H]
\centering
\begin{subfigure}[t]{0.49\textwidth}
  \includegraphics[width=\linewidth]{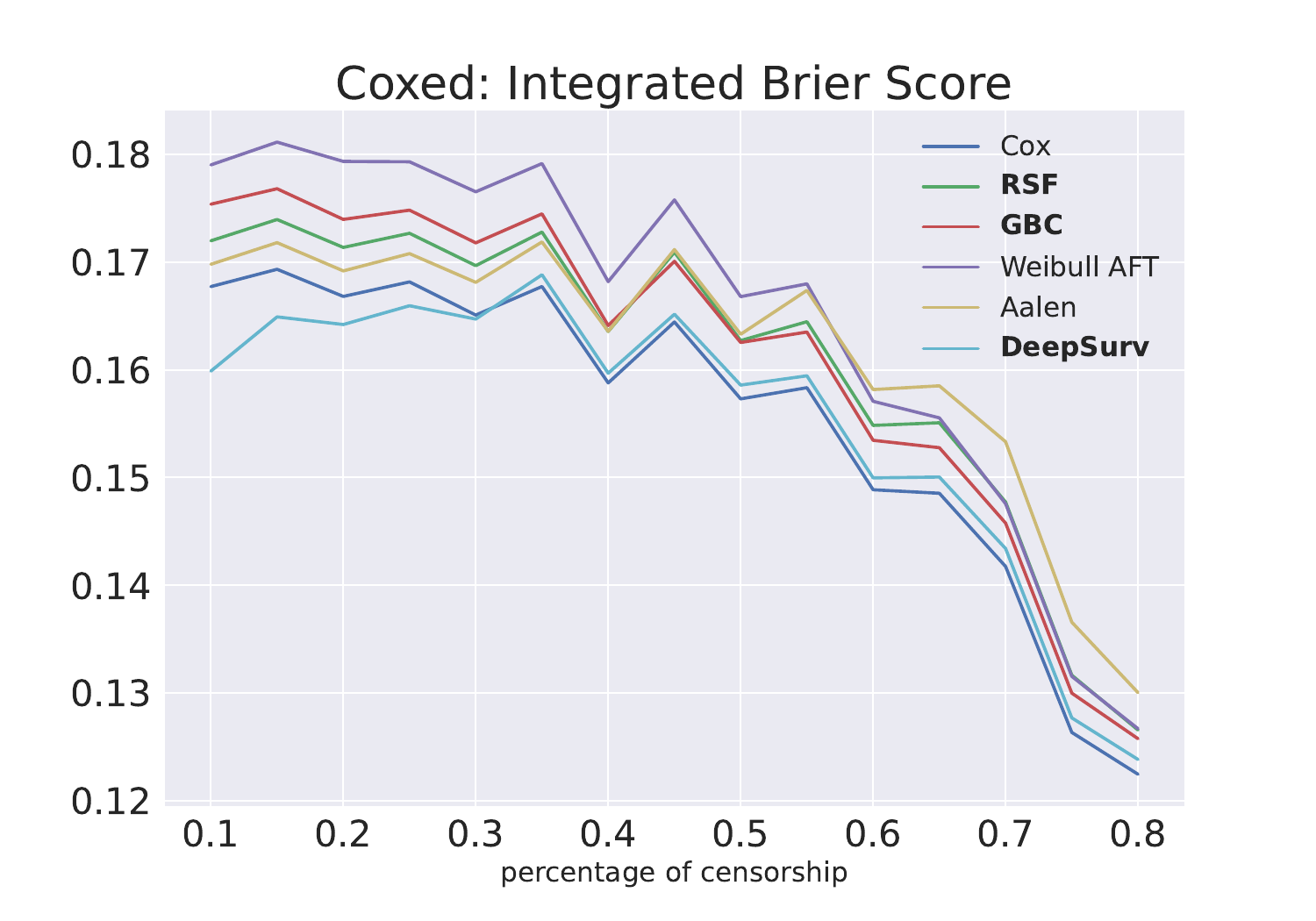}
  \caption{Integrated Brier score comparison of the increasing percentage of censorship simulation with Coxed library}
  \label{fig:coxed_pcibs}
\end{subfigure}
\hfill
\begin{subfigure}[t]{0.49\textwidth}
  \includegraphics[width=\linewidth]{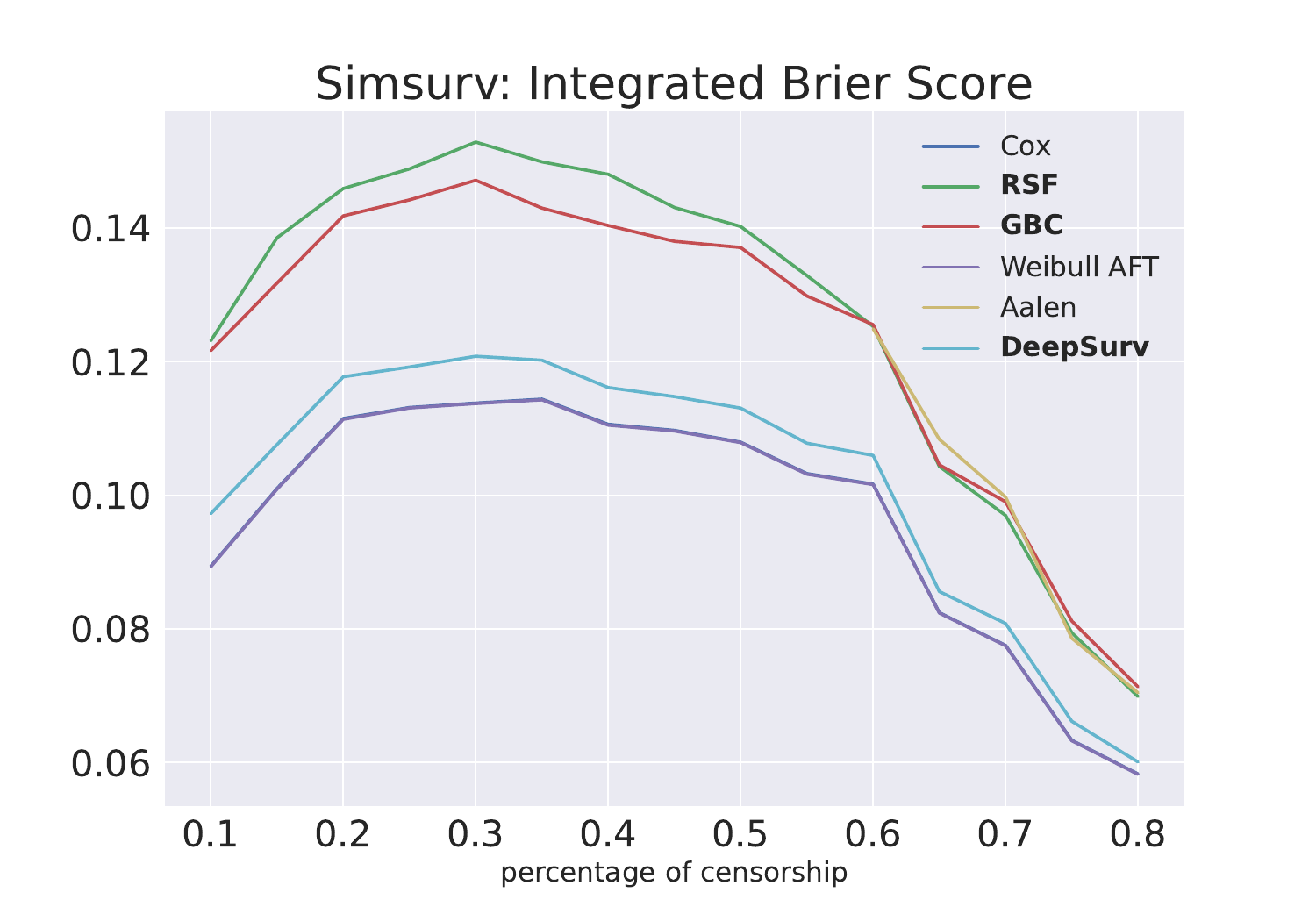}
  \caption{Integrated Brier score comparison of the increasing percentage of censorship simulation with Simsurv library}
  \label{fig:simsurv_pcibs}
\end{subfigure}
\caption{}
\end{figure}

\begin{figure}[H]
\centering
 \includegraphics[width = 0.49\linewidth]{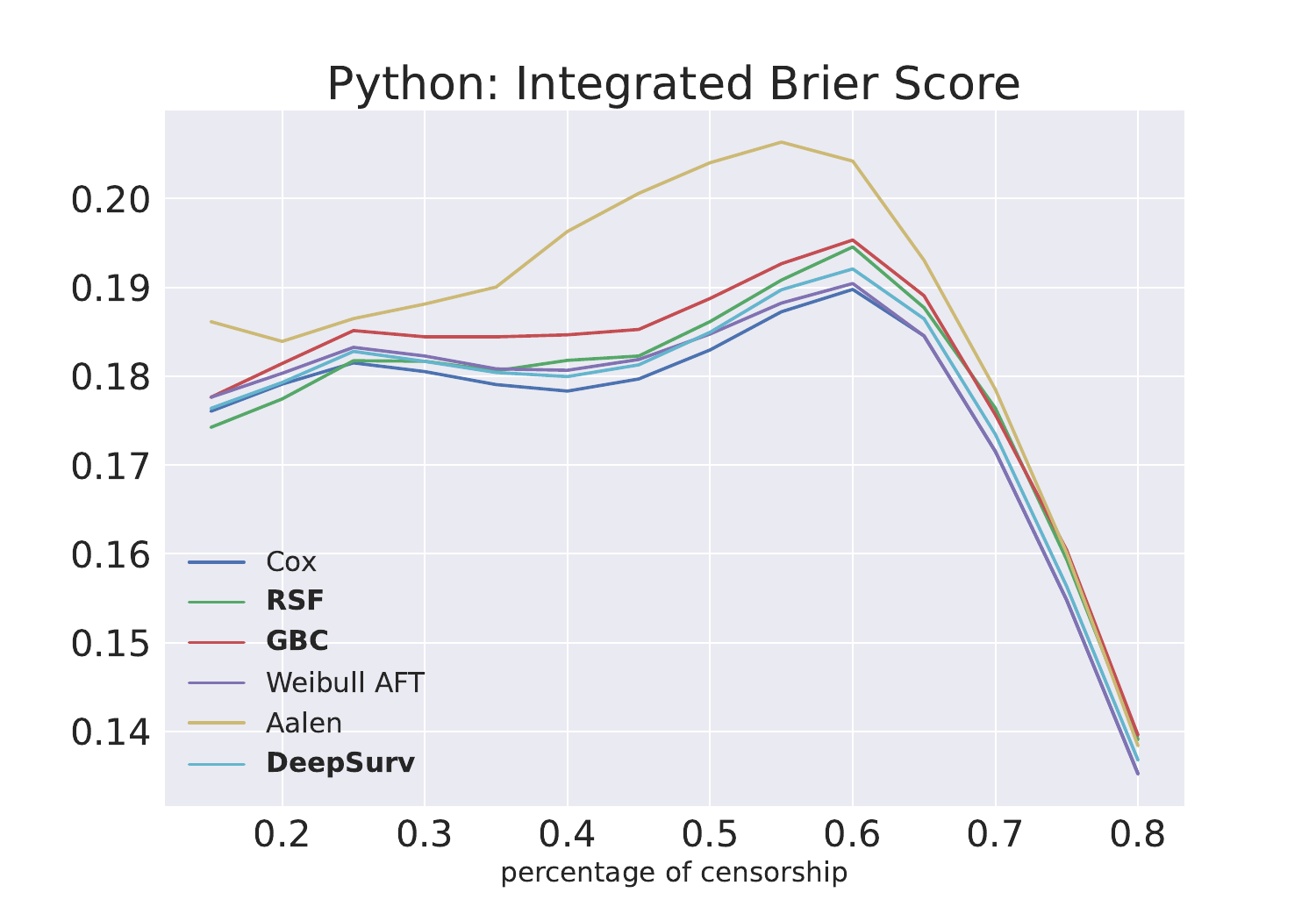}
  \caption{Integrated Brier score comparison of the increasing percentage of censorship simulation with Python}
  \label{fig:python_pcibs}
\end{figure}

\noindent
We observe in Figures \ref{fig:simsurv_pcibs} and \ref{fig:python_pcibs} an irregular increase in IBS up to $30\%$ and $60\%$ of censorship, respectively. This corresponds to the intuitive expectation that higher censorship rates should lead to poorer performance. However, when the percentage of censorship is high, as can also be seen across the entire curve in Figure \ref{fig:coxed_pcibs}, we observe an improvement in performance. This phenomenon might be attributed to the distribution of censorship. Censored individuals contribute to the Brier score only until their observed time. Therefore, if their observed time occurs at the beginning of the observation period, their contribution to the score is minimal. Consequently, if there is a significant percentage of censorship, the Brier score risks being small. Additionally, we observe that there is no significant variation in the hierarchy of the models. In Figure \ref{fig:coxed_pcibs}, Cox PH and DeepSurv consistently maintain the lead throughout the entire experiment, while in Figures \ref{fig:simsurv_pcibs} and \ref{fig:python_pcibs}, Cox PH and Weibull AFT remain at the forefront. This corroborates the conclusion drawn in the previous sections, where we found that the main factor determining the ranking of performance is the underlying distribution of the data, rather than the size of the dataset or the percentage of censorship.

\end{document}